%% file: ex_article.tex
\documentclass[final,hidelinks,onefignum,onetabnum]{siamart220329}

\usepackage{adjustbox}
\usepackage{array}
\usepackage{tabularray}

\usepackage[normalem]{ulem} 
\usepackage{todonotes}
\usepackage{graphicx} 
\usepackage{amsmath}
\usepackage{float}
\usepackage{algpseudocode}

\usepackage[font=small,labelfont=bf]{caption} 
\usepackage[subrefformat=parens]{subcaption}
\usepackage{mathtools} 
\usepackage[utf8]{inputenc} 
\usepackage[T1]{fontenc}    
\usepackage{lmodern}        
\usepackage{hyperref}       
\usepackage{url}            
\usepackage{booktabs}       
\usepackage{amsfonts}       
\usepackage{nicefrac}       
\usepackage{microtype}      
\usepackage{xcolor}         
\usepackage{wrapfig}
\usepackage{import}

\definecolor{orange_1}{HTML}{FD7F02}
\definecolor{maroon}{HTML}{C00000}
\definecolor{green}{HTML}{00AF50}
\definecolor{blue}{HTML}{0070BE}
\definecolor{red}{HTML}{FB0100}
\definecolor{yellow}{HTML}{FFC000}
\definecolor{purple}{HTML}{90408E}
\definecolor{skyblue}{HTML}{00B0F0}
\definecolor{orange}{HTML}{AE5516}

\usepackage{apptools}
\AtAppendix{\counterwithin{theorem}{section}}
\AtAppendix{\counterwithin{algorithm}{section}}
\AtAppendix{\counterwithin{table}{section}}
\AtAppendix{\counterwithin{figure}{section}}

\newtheorem{defn}{Definition}

\newcommand{\fnorm}[1]{||#1||_{\mathcal{L}_\infty}}

\allowdisplaybreaks

\input{ex_shared}

\ifpdf
\hypersetup{
  pdftitle={Unraveling architectures of continuous-time neural networks},
  pdfauthor={C. Datar, A. Datar, F.Dietrich, and W. Schilders}
}
\fi

\usepackage{xr}
\usepackage{hyperref}

\begin{document}

\maketitle

\begin{abstract}
Discovering a suitable neural network architecture for modeling complex dynamical systems poses a formidable challenge, often involving extensive trial and error and navigation through a high-dimensional hyper-parameter space. 
In this paper, we discuss a systematic approach to constructing neural architectures for modeling a subclass of dynamical systems, namely, Linear Time-Invariant (LTI) systems.
We use a variant of continuous-time neural networks in which the output of each neuron evolves continuously as a solution of a first-order or second-order Ordinary Differential Equation (ODE). 
Instead of deriving the network architecture and parameters from data, we propose a gradient-free algorithm to compute sparse architecture and network parameters directly from the given LTI system, leveraging its properties.
We bring forth a novel neural architecture paradigm featuring horizontal hidden layers and provide insights into why employing conventional neural architectures with vertical hidden layers may not be favorable.
We also provide an upper bound on the numerical errors of our neural networks. 
Finally, we demonstrate the high accuracy of our constructed networks on three numerical examples.
\end{abstract}


\import{sections/}{introduction}

\import{sections/}{dynn}
\import{sections/}{lti_preprocessing}
\import{sections/}{mapping}
\import{sections/}{algorithm}
\import{sections/}{numerical_analysis}
\import{sections/}{numerical_results}
\import{sections/}{conclusions_discussion}
\section*{Acknowledgments}
We would like to acknowledge many helpful discussions and feedback on the manuscript from Zahra Monfared, Rahul Manavalan, Iryna Burak, Erik Bolager, Ana Cukarska, Karan Shah, and Qing Sun.
While preparing this work, the authors used Grammarly to polish written text for spelling, grammar, and general style. 
\bibliographystyle{siamplain}
\bibliography{ex_article} 
\clearpage
\appendix
\import{sections/}{appendix}


\end{document}

%% file: ex_shared.tex
\usepackage{adjustbox}
\usepackage{array}
\usepackage{tabularray}

\usepackage[normalem]{ulem} 
\usepackage{algpseudocode}

\usepackage{todonotes}
\usepackage{graphicx} 
\usepackage{amsmath}
\usepackage{float}

\usepackage[font=small,labelfont=bf]{caption} 
\usepackage[subrefformat=parens]{subcaption}
\usepackage{mathtools} 
\usepackage[utf8]{inputenc} 
\usepackage[T1]{fontenc}    
\usepackage{hyperref}       
\usepackage{url}            
\usepackage{booktabs}       
\usepackage{amsfonts}       
\usepackage{nicefrac}       
\usepackage{microtype}      
\usepackage{xcolor}         
\usepackage{wrapfig}
\usepackage{import}

\ifpdf
  \DeclareGraphicsExtensions{.eps,.pdf,.png,.jpg}
\else
  \DeclareGraphicsExtensions{.eps}
\fi

\newsiamremark{remark}{Remark}
\newsiamremark{hypothesis}{Hypothesis}
\crefname{hypothesis}{Hypothesis}{Hypotheses}
\newsiamthm{claim}{Claim}

\headers{Continuous-time artificial neural networks}{C. Datar, A. Datar, F. Dietrich, and W. Schilders }

\title{Systematic construction of continuous-time neural networks for linear dynamical systems \thanks{Submitted to the editors on March 24, 2024
\funding{C.D. and W.S. acknowledge funding by the Institute for Advanced Study (IAS) at the Technical University of Munich (TUM). F.D. acknowledges funding by the DFG project no. 468830823, and the association with DFG-SPP-229. W.S. also acknowledges funding from the Dutch NWO project OCENW.GROOT.2019.044.}}}

\author{Chinmay Datar \footnotemark[4] \thanks{Institute of Advanced Study, Technical University of Munich, Germany 
  (\email{chinmay.datar@tum.de}).}
\and Adwait Datar\thanks{Institute of Data-Science Foundations, Technische Universität Hamburg, Germany
  (\email{adwait.datar@tuhh.de}).}
  \and Felix Dietrich \thanks{TUM School of Computation, Information, and Technology, Germany (\email{felix.dietrich@tum.de})}
 \and Wil Schilders \footnotemark[2] \thanks{Dept. of Mathematics and Computer Science, Eindhoven University of Technology, Netherlands (\email{w.h.a.schilders@TUE.nl}).}}

\usepackage{amsopn}


%% file: sections/introduction.tex
\section{Introduction}
\label{sec:intro}
From the evolution of quantum systems to the evolution of celestial bodies, most models in science and engineering are represented as dynamical systems in the form of differential equations. 
The exploration of neural networks in learning or modeling of dynamics is an active research field \cite{chen2024learning, otto2019linearly, qin2021data, smaoui2003analyzing}, with many applications in control \cite{bottcher-2022, kumpati-1990, nakamura2021adaptive, yin2024aonn, gaby2024neural, onken2022neural, verma2024neural}, forecasting \cite{linot2023stabilized, yeo2022generative}, and adversarial robustness \cite{celledoni2023dynamical}. 
%
The conventional discrete-time Recurrent Neural Networks (RNNs) that operate iteratively and discretely on hidden states have shown substantial progress \cite{kim-2019, chimmula-2020}. 
Unlike discrete-time RNNs, continuous-time neural networks that model a continuous evolution of hidden states between observations \cite{chen-2018, hasani-2022, lanthaler-2023, meijer-1996} have also shown significant promise in modeling dynamical systems, especially given irregularly sampled and sequential data \cite{rubanova-2019, lechner-2020, gholami-2019}.
Moreover, continuous-time neural networks are easier to impose more structure and connect machine learning to classical modeling using differential equations \cite{e-2017}.
However, numerous challenges are becoming increasingly apparent. Here, we briefly introduce three challenges that emerge when constructing RNNs: during iterative training, inference, and selecting a suitable neural architecture.
\begin{figure}[ht]
\begin{center}
    \includegraphics[width=0.83\textwidth]{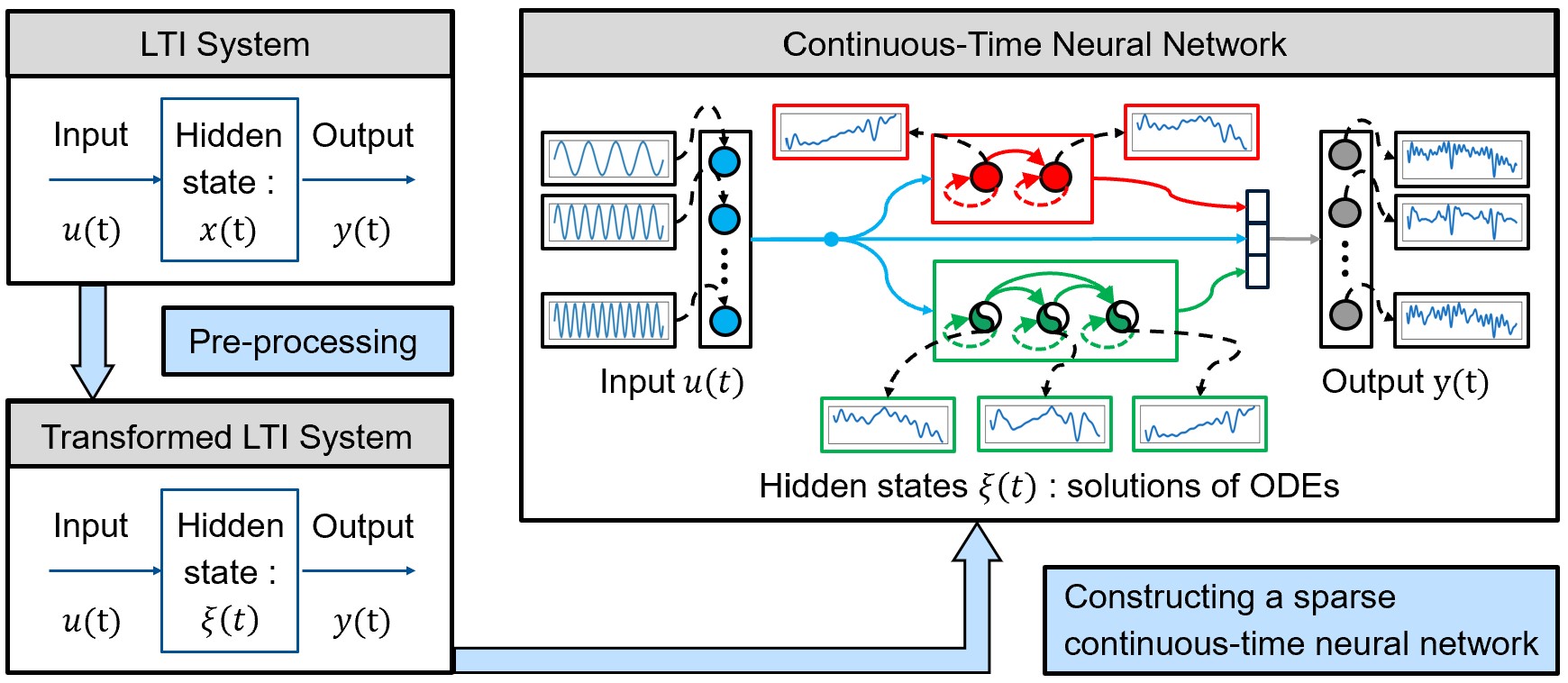}   
    \captionof{figure}{Overview of our proposed workflow illustrating the systematic construction of continuous-time neural networks from Linear Time-Invariant (LTI) systems. 
    We also showcase the architecture of our continuous-time neural network with two horizontal hidden layers (marked in red and green). 
    The states of neurons in the hidden layers are solutions of either first-order ODEs (red solid balls) or second-order ODEs (green yin-yang balls).}
    \label{fig:dynn} 
\end{center}
\vspace{-2.5em}
\end{figure}

\textbf{Challenge 1: Exploding and vanishing gradients pose a well-known challenge during iterative, gradient-based training}, especially if temporal dependencies over long intervals are present \cite{bengio-1994, mikhaeil-2022, hochreiter-2001}. 
The challenges associated with gradient-based optimization arise in discrete-time RNNs, as well as in both linear \cite{li-2021} and non-linear continuous-time neural networks \cite{meijer-1996}.
As shown in \cite{li-2021}, learning temporal relationships in data may require an exponentially large number of neurons for approximation and cause exponential slowdowns in learning dynamics -- a phenomenon described as the "curse of memory."
Motivated by the difficulties with gradient-based optimization, \cite{schilders-2009} proposes an alternative for a special class of dynamical systems, namely Linear Time-Invariant (LTI) systems (see  \cref{ss:transformations}). 
Instead of learning the network parameters from sequence data, \cite{schilders-2009} proposes using a state-space modeling algorithm \cite{verhaegen-1992, verhaegen-1993} to first identify an LTI system from data. 
A suitable architecture and network parameters are then constructed from the state-space matrices of the LTI system.
This approach provides insights into designing sparse and accurate neural networks. 
However, the work \cite{schilders-2009} is restricted to a subclass of LTI systems, namely those with the state matrix having distinct and well-separated eigenvalues. 
%
In this paper, we build upon the method presented in \cite{schilders-2009} and propose a gradient-free algorithm to construct neural networks for arbitrary LTI systems. 

\textbf{Challenge 2: Constructing sparse models is essential to speed up inference.}
Low inference times are crucial in edge computing, low-energy hardware, and applications requiring real-time response.
Even though a lot of work has been done on surrogate modeling using neural networks \cite{haasdonk2023new, tencer2021tailored, buchfink2023symplectic} for a wide range of applications such as fluid flows \cite{lui-2019, eivazi-2020, fukami-2020, li-2019}, biomechanics \cite{cicci-2023}, dynamics of mechanical systems \cite{fresca-2022}, closure modeling \cite{pan2018data}, the exact model size and capacity required for a task remain unknown. 
Empirical investigations suggest that over-parametrized models are easier to train with stochastic gradient descent \cite{kaplan-2020,brutzkus-2018,mhaskar-2016}. However, over-parameterization increases the storage requirements and computational costs associated with training and inference.
Several techniques to improve sparsity and model compression of artificial neural networks have been proposed \cite{narang-2017, choudhary-2020, hoefler2021sparsity}. 
Popular approaches include knowledge distillation \cite{hinton2015distilling}, quantization \cite{yang-2019,jacob-2018, guo-2018} or pruning \cite{anwar-2017, molchanov-2017,molchanov-2019,vadera-2022,yeom-2021,yu-2018}. 
An underlying mathematical model is unavailable for all these approaches and cannot be used for sparsification. 
In this work, we propose a pre-processing algorithm that transforms the LTI system into a form that facilitates the construction of sparse neural networks using the properties of the state matrix. 

\textbf{Challenge 3: Finding an appropriate neural network architecture} involves extensive experimentation with a lot of trial and error and dealing with a high dimensional hyper-parameter space. 
Most of the approaches in Neural Architecture Search (NAS) \cite{elsken-2019, jin-2019, liu-2018, pham-2018, ren-2021, tan-2019, wistuba-2019, zoph-2016} aim at efficiently exploring the search space of potential architectures.  
Several ways of introducing inductive biases in the model design, such as equivariance, invariance, symmetries, and recurrence, have been proposed \cite{karniadakis-2021}. 
However, these approaches still involve extensive trial and error and the exploration of numerous architectures, which often entail significant computational costs. 
In this work, we address the following question: Given a mathematical model (in our case, an LTI system), can one use it to directly compute neural network architecture and its parameters? 
We show that the properties of a given LTI system can be used to construct a sparse neural network architecture with a specific topology. 
\cref{fig:dynn} illustrates the key components of our approach. 
We explain how the concept of horizontal layers shown in \cref{fig:dynn} comes naturally in our work (see \cref{ss:mapping}). 
The key contributions of this work are as follows:
\begin{enumerate}
    \item We propose \cref{alg:preprocessing_lti} to pre-process the given LTI system with a well-conditioned transformation matrix and \cref{alg:construct_dynamic_neural_network} to construct a sparse neural network using properties of the given LTI system. 
    \item We derive a mapping from parameters of the LTI system (state-space matrices) to parameters of the neural network such that the input-output map is preserved (see \cref{th:parameter_maps}).
    \item We give an upper bound on the numerical error introduced by our neural networks (see \cref{th:numerical_analysis}).
    \item We empirically demonstrate that neural networks constructed with the proposed algorithm can simulate LTI systems accurately (see \cref{sec:numerical_results}).
\end{enumerate}

A natural question arises at this point: why model LTI systems using neural networks? 
We view this work as a first step towards constructing appropriate neural network architectures for complex dynamical systems. 
Our objective in this paper is to initiate the mathematical exploration of constructing sparse and accurate neural network models in a relatively simple and well-understood setting while systematically selecting the number of neurons, layers, and topology.
Thus, LTI systems are a good starting point for gaining insights into the network construction process. 
We emphasize that the goal of this work is not to compete with existing numerical solvers for simulating LTI systems. 



In \cref{sec:construction_lti}, we introduce LTI systems, our continuous-time neural networks, and present all the theoretical results. 
We discuss numerical experiments using our neural networks in \cref{sec:numerical_results}. 
We discuss the significance of this work, limitations, and potential extensions in \cref{sec:conclusion}. 
We now describe the notation used in this paper. 
\subsection*{Notation}
\begin{itemize}
    \item We say that a function $f(h) = \mathcal{O}(g(h))$ as $h \to 0$ if $\lim_{h \to 0} \sup \left( \frac{f(h)}{g(h)} \right)$ is finite. 
    \item We denote the space of $k$ times continuously differentiable functions over the time domain $\Omega \subset \mathbb{R}$ by $\mathcal{C}^k(\Omega)$,
    and the first-order and second-order time derivatives of functions $u \in \mathcal{C}^1(\Omega)$ and $v \in \mathcal{C}^2(\Omega)$ by $\dot{u}$ and $\ddot{v}$, respectively.
    \item We denote the identity matrix of dimensions $k \times k$ by $\mathcal{I}_k$. 
    \item For vector-valued functions $f\in\mathcal{C}(\Omega,\mathbb{R}^d)$, we write $f\in\mathcal{C}(\Omega)^d$ instead. 
    \item For a vector-valued function $u \in \mathcal{C}(\Omega)^{d \times 1}$, we write $u(t) \in \mathbb{R}^{d \times 1}$, and define the function $u^T$ such that $u^T(t) = [u(t)]^T \in \mathbb{R}^{1 \times d}$. 
    \item For any matrix $W$, we use the notation $W_{ij}$ to denote the element in the row $i$ and column $j$ of the matrix. It also denotes a block matrix represented by a block row $i$ and block column $j$.
    \item We denote the vector $\infty$-norm defined for $y\in \mathbb{R}^d$ as $||y|| =  \max_{i\in\{1,\cdots,d\}}|y_i|$.
    \item For a function $f \in \mathcal{C}(\Omega)^d$, let the $\mathcal{L}_{\infty}$ norm be $\fnorm{f} \coloneqq \sup_{x \in \Omega} ||f(x)||_{\infty}$. 
    \item We denote the state-space model of a given LTI system by matrices $\tilde{A}, \tilde{B}, \tilde{C}, \tilde{D}$ and the transformed LTI system (see \cref{alg:preprocessing_lti}) by matrices $A, B, C, D$.
\end{itemize}

%% file: sections/dynn.tex
\section{Constructing Dynamic Neural Networks for LTI Systems} \label{sec:construction_lti}
We now describe Dynamic Neural Networks (DyNNs)~\cite{meijer-1996} as a variant of continuous-time neural networks. 
We propose a pre-processing algorithm to construct new state coordinates representing a given LTI system, which helps us construct a sparse DyNN.  
We derive a mapping from the parameters of the pre-processed LTI system to the parameters of the DyNN and explain how the sparsity pattern of the transformed state matrix unravels the network architecture. 
We then discuss two algorithms: one for computing parameters and a sparse architecture of the DyNN and the other for performing a forward pass of the network to compute the output of the DyNN. 
At the end of the section, 
we derive an upper bound on the numerical error in the DyNN output. 

\subsection{Dynamic Neural Network (DyNN)} \label{ss:dynn}
\begin{figure}
    \begin{center}
    \includegraphics[width=0.67\textwidth]{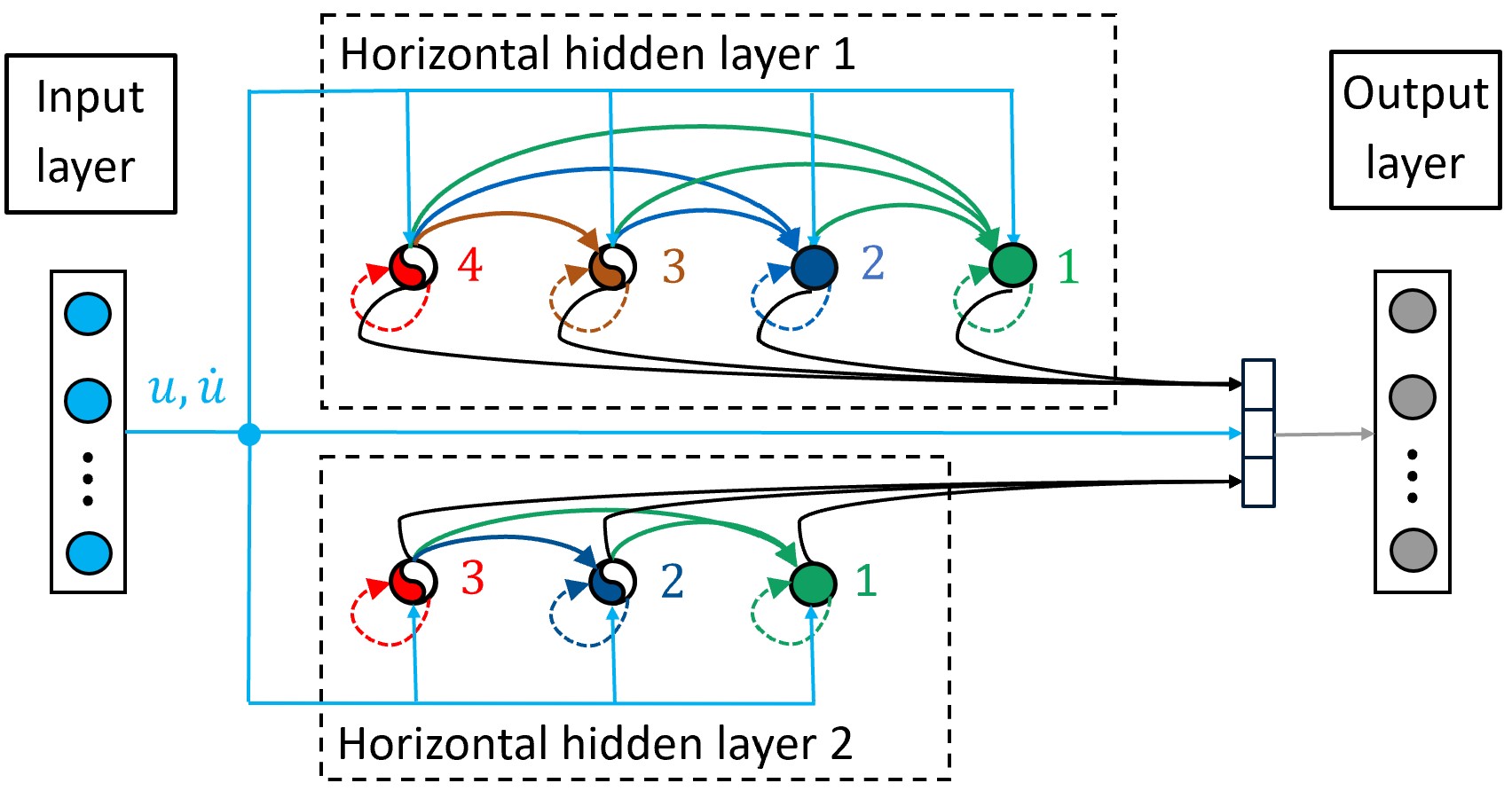}  
    \captionof{figure}{Dynamic neural network architecture with two horizontal hidden layers:   
    All neurons in the hidden layer are connected to the input layer and process inputs ($u, \dot{u}$). 
    Solid and yin-yang balls in the hidden layers indicate first-order and second-order neurons, respectively.
    The output layer is linear and is connected to all neurons in the hidden and input layers.
    The dashed self-connections of neurons in hidden layers indicate internal state dynamics. 
    }  
    \label{fig:dynn_revised}
\end{center}
\vspace{-2.5em}
\end{figure}

%
A dynamic neural network is an operator that takes a vector-valued function as input and produces a vector-valued function as output. 
In typical neural network architectures, the neurons inside ``vertical'' hidden layers are not connected to each other. 
In contrast to this, we explain the natural occurrence of horizontal layers in our work (see \cref{ss:mapping}) and demonstrate why they are necessary using a numerical example (see \cref{ss:horizontal_layers}). 
We begin by defining a dynamic neural network consisting of ``horizontal'' layers, in which the neurons within the same hidden layer have connections as shown in \cref{fig:dynn_revised}. 
Note that the neurons in the DyNN in different horizontal layers are not connected.
The input and output layers of the DyNN are not horizontal.
The output of each neuron in a horizontal layer is a solution of a first-order or second-order ODE. 
We now formally introduce DyNNs, starting with the input-output maps of neurons in the hidden layers.
\begin{defn}[Input-output map of a neuron] \label{def_io_neuron}
\textbf{The input-output map of a neuron} $i$ in hidden layer $l$ with $d_i^{(l)}$ inputs is a map $f_i^{(l)}:\mathcal{C}(\Omega)^{d_i^{(l)}} \ni u^{(l)}_i \mapsto y_i^{(l)}\in \mathcal{C}^1(\Omega)^2$ defined via the solution to the differential equation 
\begin{align*}
m^{(l)}_i \; \ddot{\xi}^{(l)}_{i}(t) + c^{(l)}_i \; \dot{\xi}^{(l)}_{i}(t) + k^{(l)}_i\xi^{(l)}_{i}(t) &=
w_i^{(l)} u_i^{(l)}(t), \quad \xi_i^{(l)}(0)=0,\quad \dot{\xi}_i^{(l)}(0)=0,\\
y_i^{(l)}(t)&=\begin{bmatrix}
    \xi_i^{(l)}(t) & \dot{\xi}_i^{(l)}(t)
\end{bmatrix}^T,
\end{align*}
where $w_i^{(l)}\in \mathbb{R}^{1\times d_i^{(l)}}$, $m^{(l)}_i,c^{(l)}_i,k^{(l)}_i \in \mathbb{R}$ are the weights of the neuron, $\xi^{(l)}_{i}$ is the state of the neuron. We say that the map $f_i^{(l)}$ is defined corresponding to $(m^{(l)}_i, c^{(l)}_i,k^{(l)}_i, w_i^{(l)})$. 
Furthermore, if $m_i^{(l)}=0$, we refer to the neuron as a \textbf{``first-order neuron''}. Otherwise, it is called a \textbf{``second-order neuron''}.
\end{defn}

The neural network architecture describing how neurons are interconnected with each other is shown in \cref{fig:dynn_revised} and is described next.
Let $n_l$ be the number of neurons in the horizontal layer $l$. 
Let $d_i$ and $d_o$ be the number of neurons in the input and output layers of the DyNN, respectively.
The architecture shown in \cref{fig:dynn_revised} can be described by the input-output map of each neuron as 
\begin{equation} \label{eq:neuron_map}
    y_i^{(l)}=f_i^{(l)}(u_i^{(l)}), \, \text{where} \  
    u_i^{(l)}=\begin{bmatrix}
        u^T& \dot{u}^T & [y_{i+1}^{(l)}]^T &[y_{i+2}^{(l)}]^T&\cdots&[y_{n_l}^{(l)}]^T
    \end{bmatrix}^T
\end{equation}
for $i \in \{1,\cdots,n_l\}$ and $l\in\{1,\cdots,L\}$.
The topology of the neurons in the horizontal hidden layers of the network implies that a neuron $i$ in horizontal layer $l$ has the input dimension $d_i^{(l)} = 2d_i + 2(n_l-i)$, where the term $2d_i$ stems from inputs $u$ and $\dot{u}$ and $2(n_l-i)$ from states of other neurons in the same hidden layer and their derivatives.
The hidden layers of a DyNN essentially represent a coupled system of ODEs whose parameters are $(m^{(l)}_i,c^{(l)}_i,k^{(l)}_i, w_i^{(l)})$. 
First-order neurons generally model state dynamics without oscillations, whereas second-order neurons model state dynamics with oscillations.
The output layer of a dynamic neural network is a linear layer with connections from all neurons in the hidden and input layers with parameters $\phi_i^{(l)} \in \mathbb{R}^{d_o \times 2}$ and $\Psi \in \mathbb{R}^{d_o \times d_i}$, respectively. 
We next define suitable sets of parameters and input-state-output maps of the dynamic neural network, which facilitate the discussion in \cref{ss:mapping}.
\begin{defn}[Parameter sets of the dynamic neural network] \label{eq:parameter_sets_dynn}
   For positive integers $d_i$, $d_o$, $L$, $n_l$ and $l \in \{1, \cdots, L\}$, let the tuple of weights $m_i^{(l)}$ in layer $l$ be  
    \begin{align*}
        \mathcal{M}^{(l)}&=\left(m^{(l)}_1,\cdots,m^{(l)}_{n_l}\right) \quad \textnormal{and let} \quad \mathcal{M}=\left(\mathcal{M}^{(1)},\cdots,\mathcal{M}^{(L)}\right).
    \end{align*}
    We analogously define $\mathcal{C}^{(l)},\mathcal{C}$ as a collection of all $c_i^{(l)}$; $\mathcal{K}^{(l)},\mathcal{K}$ of all $k_i^{(l)}$; $\mathcal{W}^{(l)},\mathcal{W}$ of all $w_i^{(l)}$ and $\Phi^{(l)},\Phi$ of all $\phi_i^{(l)}$.
    Finally, define the sets 
    \begin{align*}
    \mathcal{P}_{dynn}^{(l)}&:=\left\{
    \left(\mathcal{M}^{(l)},\mathcal{C}^{(l)},\mathcal{K}^{(l)},\mathcal{W}^{(l)}\right): m^{(l)}_i,c^{(l)}_i,k^{(l)}_i\in \mathbb{R}, w_i^{(l)}\in \mathbb{R}^{1\times d_i^{(l)}}, 1 \leq i \leq n_l \right\}, \\
    \mathcal{P}_{dynn}^{hidden}&:=\left\{
    \left(\mathcal{M},\mathcal{C},\mathcal{K},\mathcal{W}\right): \left(\mathcal{M}^{(l)},\mathcal{C}^{(l)},\mathcal{K}^{(l)},\mathcal{W}^{(l)}\right) \in \mathcal{P}_{dynn}^{(l)}, 1 \leq l \leq L
    \right\},\\
    \mathcal{P}_{dynn}^{output}&:=\left\{
    \left(\Phi,\Psi\right):           
        \phi_i^{(l)} \in \mathbb{R}^{d_o \times 2},\Psi \in \mathbb{R}^{d_o \times d_i}, 1 \leq i \leq n_l, 1 \leq l \leq L \right\},
    \end{align*}

    which collect all parameters of the hidden layer $l$, all parameters of all hidden layers, and all parameters of the output layer, respectively.
\end{defn}

\begin{defn}[Input-state-output maps of DyNN]\label{def_iso_dynn}
Consider a dynamic neural network with $L$ horizontal layers, $n_l$ neurons in the horizontal layer $l$, $d_i$ neurons in the input layer, and $d_o$ neurons in the output layer. 
Let $\left(\mathcal{M},\mathcal{C},\mathcal{K},\mathcal{W}\right) \in \mathcal{P}_{dynn}^{hidden}$ and $\left(\Phi,\Psi\right) \in \mathcal{P}_{dynn}^{output}$ be the parameters of the DyNN.
The forward pass of the DyNN for an arbitrary input $u \in \mathcal{C}^1(\Omega)^{d_i}$ can be described by
\begin{align} \label{eq:f_dynn}
    y(t)&= \biggl(\sum_{l=1}^{L} \sum_{i=1}^{n_l} \phi_i^{(l)}  \, y_i^{(l)}(t) \biggr) + \Psi \, u(t),\quad \textnormal{ for } \, t \in \Omega,\\
    y_i^{(l)}(t)&=\begin{bmatrix}
    \xi_i^{(l)}(t)&\dot{\xi}_i^{(l)}(t)
\end{bmatrix}^T=f_i^{(l)}(u_i^{(l)})(t),\\
    u_i^{(l)}(t)&=\begin{bmatrix}
        u^T(t)&\dot{u}^T(t)&[y_{i+1}^{(l)}(t)]^T&[y_{i+2}^{(l)}(t)]^T&\cdots&[y_{n_l}^{(l)}(t)]^T
    \end{bmatrix}^T,     
\end{align}
where $f_i^{(l)}$ is the input-output map corresponding to $(m^{(l)}_i,c^{(l)}_i,k^{(l)}_i, w_i^{(l)})$ described in \cref{def_io_neuron}.
Based on these equations, the \textbf{input-output map of DyNN} $f_{dynn}:\mathcal{C}^1(\Omega)^{d_i} \rightarrow \mathcal{C}^1(\Omega)^{d_o}$, the \textbf{input-state map of DyNN} $f_{dynn}^s:\mathcal{C}^1(\Omega)^{d_i} \rightarrow \mathcal{C}^1(\Omega)^{(n_1+\cdots+n_L)}$ and the \textbf{input-state map of the $l^{\textnormal{th}}$ hidden layer of DyNN} $f_{dynn}^{(l)}:\mathcal{C}^1(\Omega)^{d_i} \rightarrow \mathcal{C}^1(\Omega)^{n_l}$ are defined as
\begin{align*}
    f_{dynn}&:u\mapsto y, \quad 
    f_{dynn}^s:u\mapsto \xi, \textnormal{ where }\xi(t)=\begin{bmatrix}[\xi^{(1)}(t)]^T&\hdots& [\xi^{(L)}(t)]^T\end{bmatrix}^T,\\
    f_{dynn}^{(l)}&:u\mapsto \xi^{(l)}, \textnormal{ where }\xi^{(l)}(t)=\begin{bmatrix}\xi^{(l)}_1(t)&\hdots& \xi^{(l)}_{n_l}(t)\end{bmatrix}^T, \quad l\in \{1,\cdots,L\}.
\end{align*}
We call $\xi^{(l)}$ the \textbf{state of the horizontal layer $\mathbf{l}$}, and $\xi$ the \textbf{state of the DyNN}.
\end{defn}

%% file: sections/lti_preprocessing.tex
\subsection{LTI systems and pre-processing} \label{ss:transformations}
For positive integers $d_i$, $d_h$, $d_o$, all LTI systems are determined by four matrices: state matrix $\tilde{A} \in \mathbb{R}^{d_{h} \times d_{h}}$, input matrix $\tilde{B} \in \mathbb{R}^{d_{h} \times d_{i}}$, output matrix $\tilde{C} \in \mathbb{R}^{d_{o} \times d_{h}}$, and feed-forward matrix $\tilde{D} \in \mathbb{R}^{d_{o} \times d_{i}}$. 
We transform a given LTI system into a sparse representation, which then facilitates constructing a sparse dynamic neural network. 
The state-space representation of a general LTI system is 
\begin{subequations}\label{eq:lti_0}
    \begin{align} \label{eq:initial_state_eq}
        \dot{x}(t) &=  \tilde{A} \; x(t) \;+\;  \tilde{B} \; u(t), \quad \quad x(0) = 0,\\
   \label{eq:initial_output_eq}
        y(t) &=\tilde{C} \; x(t) \;+\; \tilde{D} \; u(t),
    \end{align}
\end{subequations}
where, at time $t$, $x(t) \in \mathbb{R}^{d_{h}}$ is the state, $ u(t) \in \mathbb{R}^{d_{i}}$ is the input, and $ y(t) \in \mathbb{R}^{d_{o}} $ is the output of the system. 
Here, linearity means that the map $u \mapsto y $ is linear, and time-invariance means that the state-space matrices are independent of time.

We represent the state variable $x(t)$ in the state-space formulation
by the hidden state of neurons in a dynamic neural network. 
Thus, the sparsity pattern of the state matrix $\tilde{A}$ determines the topology and number of connections in hidden layers of the dynamic neural network that we use to model the LTI system. 
For a dense state matrix $\tilde{A}$, neurons in hidden layers of a dynamic neural network would result in a fully connected graph with many neural connections. 
Given any LTI system, we propose the pre-processing \cref{alg:preprocessing_lti} to block-diagonalize the state matrix $\tilde{A}$ using similarity transformations with well-conditioned transformation matrices. 
This preserves the input-output map from $u$ to $y$, making the transformed state matrix $\tilde{A}$ sparse.

\newcommand{\comment}[2][.6\linewidth]{%
  \leavevmode\hfill\makebox[#1][l]{//~#2}}
\algnewcommand\algorithmicto{\textbf{to}}
\algnewcommand\RETURN{\State \textbf{return} }
\begin{algorithm}[ht]
    \caption{Pre-processing the LTI system}\label{alg:preprocessing_lti}
    \textbf{Input:}  State Space Matrices($\tilde{A}, \tilde{B}, \tilde{C}, \tilde{D}$)\\
    \textbf{Output:} Transformed State Space Matrices ($A, B, C, D$)\\
    \textbf{Parameters:} \texttt{clustering algorithm}
    \begin{algorithmic}[1]
        \State $\mathcal{R} \gets \mathcal{T}_1^T \tilde{A} \mathcal{T}_1$ \comment{Real Schur decomposition }
        \If{$\mathcal{R}$ is diagonal} \comment{If $\tilde{A}$ is unitarily diagonalizable}
            \State $\mathcal{T} = \mathcal{T}_1$ 
            \comment{Transformation matrix}
            \State $A \gets \mathcal{R}$ 
        \ElsIf{$\mathcal{R}$ is not diagonal} \comment{If $\tilde{A}$ is not unitarily diagonalizable}
            \State $\tilde{\mathcal{R}} \gets \mathcal{T}_2^T \tilde{A} \mathcal{T}_2$
            \comment{Ordered real Schur form (\cite{bai1993swapping}) using $\mathcal{R}$}
            \State $A \gets \mathcal{T}_3^{-1} \mathcal{\tilde{R}}  \mathcal{T}_3$ \comment{Block-diagonalization (Bartels-Stewart \cite{bartels-1972})}
            \State  $\mathcal{T} = \mathcal{T}_2 \mathcal{T}_3$ \comment{Transformation matrix}
        \EndIf
        \State  ($B$, $C$, $D$) $\gets$ ($\mathcal{T}^{-1} \tilde{B}, \tilde{C}\mathcal{T}, \tilde{D}$)
        \comment{New State Space Matrices}
    \end{algorithmic}
\end{algorithm} 
In the first step of Algorithm \ref{alg:preprocessing_lti}, we perform real Schur decomposition given by $\mathcal{R} = \mathcal{T}_1^T \tilde{A} \mathcal{T}_1$, 
where $\mathcal{T}_1$ is an orthogonal matrix, and  $\mathcal{R}$ is a block-upper-triangular matrix with diagonal blocks of dimensions $1 \times 1$ and $2 \times 2$ corresponding to real and pairs of complex eigenvalues, respectively. 
If the matrix $\tilde{A}$ is unitarily diagonalizable, $\mathcal{R}$ is a diagonal matrix. 
This is ideal and results in the sparsest possible state matrix. 
If $\tilde{A}$ is not unitarily diagonalizable, we proceed with a modified version of the ordered real Schur decomposition proposed in \cite{bai1993swapping} to order the eigenvalues appearing on the diagonal of the state matrix. 
We use the \texttt{clustering algorithm} (see \cref{rem:sm_clustering} for details) to cluster eigenvalues in 
$L$ clusters, where $L$ is a hyper-parameter. 
If $\tilde{A}$ is not unitarily diagonalizable, it is important to choose $L$ so that eigenvalues in distinct clusters are sufficiently apart.
We specify the order of eigenvalues so that eigenvalues from each cluster appear on the diagonal sequentially, one cluster after the other (see \cref{rem:sm_clustering} for details on ordering) by modifying the implementation \cite{githubPythonCode}. 
Thus, the transformation $\tilde{\mathcal{R}} \gets \mathcal{T}_2^T \tilde{A} \mathcal{T}_2$ reduces the state matrix to a block-upper triangular matrix with $L$ diagonal blocks, with each block corresponding to one cluster of eigenvalues. 
This condition is necessary to apply the Bartels-Stewart algorithm to avoid ill-conditioned transformation matrices. 
The Bartels-Stewart algorithm is a similarity transformation $\mathcal{T}_3^{-1} \mathcal{\tilde{R}} \mathcal{T}_3$ that reduces all the off-diagonal entries of $\mathcal{\tilde{R}}$ to zero and block-diagonalizes the state matrix. 
The transformation of a non-unitarily-diagonalizable state matrix could be summarized (for $\tilde{A}_{ij}$ representing an element in row $i$ and column $j$, $\tilde{R}_{ij}$ representing block-row $i$ and block-row $j$) as: 
\begin{equation} \label{eq:transformations}
\begin{gathered}
    \underbrace{
    \begin{bmatrix}
    \tilde{A}_{11} & \hdots & \tilde{A}_{1 d_h} \\
    \vdots &  \ddots & \vdots \\
    \tilde{A}_{d_h 1} & \hdots & \tilde{A}_{d_h d_h} \\
    \end{bmatrix}
    }_{\tilde{A}}\hspace{-0.2em}
    \xRightarrow{\mathcal{T}_2^T \tilde{A} \mathcal{T}_2}
    \hspace{-0.2em}
    \underbrace{
    \begin{bmatrix}
    \tilde{\mathcal{R}}_{11} &  \hdots & \tilde{\mathcal{R}}_{1L} \\
                             &  \ddots & \vdots \\    
                             &         & \tilde{\mathcal{R}}_{LL} \\
    \end{bmatrix}
    }_{\tilde{\mathcal{R}}}\hspace{-0.2em}
    \xRightarrow{\mathcal{T}_3^{-1} 
    \hspace{-0.2em}
    \tilde{\mathcal{R}} \mathcal{T}_3}
     \underbrace{
    \begin{bmatrix}
         \mathcal{\tilde{R}}_{11} &  &  \\
                &  \mathcal{\tilde{R}}_{22} & \\
                &   &  \mathcal{\tilde{R}}_{LL} \\
    \end{bmatrix}    
    }_{A}.
    \end{gathered}
\end{equation}
See \cref{app:arithmatic_complexity} for the algebraic complexity of the algorithm. 
With a specially tailored example, we will also illustrate the process of selecting $L$ and its impact on the condition number of the transformation matrix (see \cref{ss:horizontal_layers}). 

We do not transform the state matrix into other canonical forms, such as controller canonical form or Jordan canonical form, for generalizability and to avoid highly ill-conditioned transformation matrices.
After block-diagonalization of the state-matrix with \cref{alg:preprocessing_lti}, each diagonal block has either all real, all complex, or mixed eigenvalues that are systematically ordered. 
If a diagonal block has mixed eigenvalues, the real eigenvalues appear first, followed by pairs of complex eigenvalues. 
We now define sets of sparse state matrices that describe the three possible sparsity patterns of any diagonal block of $A$.
\begin{defn}[Sets of sparse state matrices]\label{def:sparse_G}
    Let ${\mathcal{G}}$ be the set of all block-upper triangular matrices $M$ such that
    (a) if $M$ has $k_r$ real eigenvalues, then all blocks in the first $k_r$ rows of $M$ are of dimension $1 \times 1$, and (b) if $M$ has $k_c$ pairs of complex eigenvalues with non-zero imaginary parts, then  all blocks in the last $2k_c$ rows of $M$ are of dimension $2 \times 2$ and these blocks have non-zero entries in the upper-right corner, i.e., if $\begin{bmatrix} a&b\\c&d \end{bmatrix}$ is a diagonal block in the last $2k_c$ rows, then $b\neq 0$. 
    Let ${\mathcal{G}_r}$ $\subset {\mathcal{G}}$ 
    and ${\mathcal{G}_c}$ $\subset {\mathcal{G}}$ 
    be the subsets containing matrices having all real eigenvalues ($k_c=0$) and all eigenvalues having non-zero imaginary parts ($k_r=0$) respectively.
\end{defn}
The sparsity patterns of the diagonal blocks, depending on whether they have real only, complex only, or mixed eigenvalues, are 
\begin{equation*} \label{eq:sparsity_patterns}
    \begin{gathered}
    \underbrace{
    \begin{bmatrix}
                    * & \hdots   & * \\
                    & \ddots & \vdots\\
                    & & * \\
    \end{bmatrix}
    }_{\mathcal{G}_r},
    \underbrace{
    \left[ 
        \begin{array}{c c c c c c c} 
             \ast & \ast & \ast & \ast & \hdots & \ast & \ast\\ 
             \ast & \ast & \ast & \ast & \hdots & \ast & \ast\\ 
              &  & \ast & \ast & \hdots & \ast & \ast\\ 
              &  & \ast & \ast & \hdots & \ast & \ast\\ 
              &  &  &  & \ddots & \vdots & \vdots \\ 
             &  &  &  &  &  \ast & \ast\\ 
              &  &  &  &  &  \ast & \ast\\
          \end{array} 
        \right]
        }_{\mathcal{G}_c},
        \underbrace{
        \left[ 
        \begin{array}{c c c c c c c c} 
            \ast & \hdots & \ast & \hdots & \hdots & \hdots & \hdots & \ast\\ 
            & \ddots & \vdots & \vdots & \vdots & \vdots & \vdots & \vdots\\ 
             & & \ast & \hdots & \hdots & \hdots & \hdots & \ast\\ 
              & &  & \ast & \ast & \hdots & \ast & \ast \\ 
              & &  & \ast & \ast & \hdots & \ast & \ast \\ 
              & &  &  &  & \ddots &  \vdots & \vdots \\ 
             &  &  &  &  & & \ast & \ast\\ 
             &  &  &  &  & & \ast & \ast\\
          \end{array} 
        \right]}_{\mathcal{G}}.
    \end{gathered}
\end{equation*}
In summary, we reduce the state matrix $\tilde{A}$ to a block-diagonal form $A$ as shown in \cref{eq:transformations}.
We now define a new coordinate system $x(t) = \mathcal{T} \xi(t)$ 
such that the transformed LTI system in the new state coordinates $\xi(t)$ is
\newline
    \begin{subequations} \label{eq:transformed_lti_sys}
    \begin{align} \label{eq:transformed_state_eq}
    \begin{bmatrix}
        \dot{\xi}^{(1)}(t) \\
        \vdots\\
        \dot{\xi}^{(L)}(t)
    \end{bmatrix}
    &= 
    \underbrace{
    \begin{bmatrix}
        A_{11} & & \\
        & \ddots & \\
        & & A_{LL}
    \end{bmatrix}}_{A}
    \begin{bmatrix}
        \xi^{(1)}(t) \\
        \vdots\\
        \xi^{(L)}(t)
    \end{bmatrix}
    +
    \underbrace{
    \begin{bmatrix}
        B^{(1)} \\
        \vdots \\
        B^{(L)}
    \end{bmatrix}}_{B}
    u(t),
    \quad \xi(0) = 0,
    \end{align}
    \begin{align} \label{eq:transformed_output_eq}
        y(t) = C \; \xi(t) \;+\; D \; u(t) \quad {\textnormal{, where }}
    \end{align}
    \begin{equation}
        A = \mathcal{T}^{-1} \tilde{A} \mathcal{T}, \quad B = \mathcal{T}^{-1} \tilde{B}, \quad C = \tilde{C}\mathcal{T}, \quad D = \tilde{D}.
    \end{equation}
\end{subequations}
Let $l \in \{1, 2, ... , L\}$. 
Let the dimensions of the diagonal block $l$ be $d_l \times d_l$. 
Let the states and time-derivatives of states in the block row $l$ be denoted by $\xi^{(l)}, \dot{\xi}^{(l)}$. 
Let the block row $l$ of the matrix $B$ be denoted by $B^{(l)}$. 
Finally, we provide the definitions concerning the parameter sets of an LTI system and the input-state-output maps of an LTI system, which are required for the discussion in Section \ref{ss:mapping}. 
\begin{defn}[Parameters of an LTI system]
    Corresponding to positive integers $d_h,d_i,d_o$, define $\mathcal{P}_{lti}^{state}:=\left\{\left(A,B\right):A\in \mathcal{S}\subset \mathbb{R}^{d_h \times d_h}, B\in \mathbb{R}^{d_h\times d_i}\right\},$ and $\mathcal{P}_{lti}^{output}:=\left\{\left(C,D\right):C\in \mathbb{R}^{d_o\times d_h},D\in\mathbb{R}^{d_o\times d_i}\right\},$
    where $\mathcal{S}$ is the set of all square matrices that are block-diagonal, with each diagonal block belonging to ${\mathcal{G}}$.
\end{defn}
\begin{defn}[Input-state-output maps of an LTI system]
\label{def:f_state}
    Let $A \in \mathbb{R}^{d_h \times d_h}, B \in \mathbb{R}^{d_h \times d_i},C\in \mathbb{R}^{d_o\times d_h},D\in\mathbb{R}^{d_o\times d_i}$.
    Let the state $x\in \mathcal{C}^1(\Omega)^{d_h},$ input $u\in \mathcal{C}(\Omega)^{d_i}$ and output $y\in \mathcal{C}(\Omega)^{d_o}$ be related by the governing equations of an LTI system
    \begin{align}\label{eq:state_equation_new}
        \dot{x}(t) &= Ax(t) + Bu(t), \quad x(0) = 0,  \\
        \label{eq:f_lti}
        y(t)&= Cx(t) + D u(t),
    \end{align}
    for $t \in \Omega$.
    For this LTI system, define the \textbf{input-state map of the LTI system} corresponding to $(A,B)$ as $f_{lti}^s:\mathcal{C}(\Omega)^{d_i}\ni u \mapsto x \in \mathcal{C}^1(\Omega)^{d_h} $ defined via \eqref{eq:state_equation_new} and the \textbf{input-output map of the LTI system} corresponding to $(A,B,C,D)$ as $f_{lti}:\mathcal{C}(\Omega)^{d_i}\ni u \mapsto y \in \mathcal{C}(\Omega)^{d_o}$ defined via \eqref{eq:state_equation_new} and \eqref{eq:f_lti}.
\end{defn}



%% file: sections/mapping.tex
\subsection{Mapping from parameters of the LTI system to parameters of the DyNN} \label{ss:mapping}
We seek to construct a dynamic neural network such that its input-output map equals the input-output map of a given LTI system. 
Note that the input-output map of our DyNN can be described via the solution of a coupled system of first-order and/or second-order ODEs.
This can be seen by assembling the ODEs corresponding to all neurons and substituting the interconnection structure (see \cref{lemma: overall_dynamics_DyNN}).
As an intermediate technical result, we show that the input-output map of an LTI system can also be represented via the solution of a coupled system of first-order and/or second-order ODEs (see \cref{lemma:m_lti_to_second_order}).
In \cref{th:parameter_maps}, which is the main result of this section, we show how to construct the parameters of a DyNN, preserving the input-output map of a given LTI system. 
See \cref{fig:theory_workflow} for a visual representation of how different theoretical results are interconnected and used in \cref{th:parameter_maps}.
We start by defining the set of tuples of matrices that describe a coupled system of ODEs represented by each horizontal layer of the DyNN. 
\begin{defn} [Tuples of matrices defining first and/or second order system] \label{def:matrices_2nd_order_formalism}
    Corresponding to $n_l, d_i \in \mathbb{N}$, let $\mathcal{S}_{n_l, d_i}$ be the set of tuples $(M, C, K, E, V)$ where $M, C, K \in \mathbb{R}^{n_l \times n_l}, E, V,  \in \mathbb{R}^{n_l \times d_i}$, $M$ is a diagonal matrix and $C$, $K$ are upper-triangular matrices.
\end{defn}
\begin{lemma}[First and/or second order dynamics of DyNN] \label{lemma: overall_dynamics_DyNN}
Consider a dynamic neural network with $L$ horizontal layers, $n_l$ neurons in the horizontal layer $l$, $d_i$ neurons in the input layer, and $d_o$ neurons in the output layer.
For $l\in\{1, \cdots, L\}$, let $\left(\mathcal{M}^{(l)},\mathcal{C}^{(l)},\mathcal{K}^{(l)},\mathcal{W}^{(l)}\right) \in \mathcal{P}_{dynn}^{(l)}$ be the parameters of hidden layers of the DyNN.
Let $u \in \mathcal{C}^{1}(\Omega)^{d_i}$ be an arbitrary input and $\xi^{(l)}$ be the state of the $l^{\textnormal{th}}$ hidden layer, i.e., $\xi^{(l)}=f_{dynn}^{(l)}(u)$.
Then for $l\in \{1,\cdots,L\}$, a bijective mapping 
\begin{equation*}
\mathfrak{n}^{(l)}_{dynn}: \mathcal{P}_{dynn}^{(l)} \ni \left(\mathcal{M}^{(l)},\mathcal{C}^{(l)},\mathcal{K}^{(l)},\mathcal{W}^{(l)}\right) \mapsto \left(M^{(l)}, C^{(l)}, K^{(l)}, E^{(l)}, V^{(l)}\right) \in \mathcal{S}_{n_l,d_i},     
\end{equation*} 
described in \cref{ssec:n_dynn} can be constructed such that $\xi^{(l)}$ solves
\begin{equation}\label{eq:sos}
    M^{(l)}\Ddot{\xi}^{(l)}(t)+C^{(l)}\dot{\xi}^{(l)}(t)+K^{(l)}\xi^{(l)}(t)=E^{(l)}u(t)+V^{(1)}\dot{u}(t) \quad \forall t \in \Omega
\end{equation}
with zero initial conditions.
Conversely, for arbitrary $\left(M^{(l)}, C^{(l)}, K^{(l)}, E^{(l)}, V^{(l)}\right)\in \mathcal{S}_{n_l,d_i}$, $l\in\{1, \cdots, L\}$, if $\xi^{(l)}$ solves the differential equation \eqref{eq:sos} with zero initial conditions, then one can construct a DyNN with parameters $\left(\mathcal{M}^{(l)},\mathcal{C}^{(l)},\mathcal{K}^{(l)},\mathcal{W}^{(l)}\right)$ computed by the inverse of $\mathfrak{n}_{dynn}^{(l)}$ such that $\xi^{(l)}=f_{dynn}^{(l)}(u)$.
\end{lemma}
\begin{proof}
    See \cref{app_lemma: overall_dynamics_DyNN}. 
\end{proof}
Depending on whether all, none, or few of the entries of $\mathcal{M}^{(l)}$ are zero, states of hidden layer $l$ of a DyNN represent a coupled linear system of solely first-order ODEs or solely second-order ODEs or a combination of both, respectively. 
\begin{theorem}[Mapping an LTI system to a DyNN] \label{th:parameter_maps}
For positive constants $d_h,d_i,d_o$, consider an LTI system defined by $\left(A,B\right)\in \mathcal{P}_{lti}^{state}$ and $\left(C,D\right)\in \mathcal{P}_{lti}^{output}$.
For positive integers $L$, $n_l$ and for $l \in \{1, \cdots, L\}$, mappings 
\begin{align*}
    \mathfrak{m}_h&:\left(A,B\right) \mapsto \left(\mathcal{M},\mathcal{C},\mathcal{K},\mathcal{W},\Theta\right) \in \mathcal{P}_{dynn}^{hidden},\\
    \mathfrak{m}_o&:\left(A,B,C,D\right) \mapsto \left(\Phi,\Psi\right) \in \mathcal{P}_{dynn}^{output},
\end{align*}
as described in \cref{ssec:m_h} and \cref{ssec:m_0}, respectively, can be constructed such that the DyNN with parameters $\left(\mathcal{M},\mathcal{C},\mathcal{K},\mathcal{W},\Theta\right)$ and $\left(\Phi,\Psi\right)$ satisfies the property that $f_{dynn}(u)=f_{lti}(u)$ for all $u \in \mathcal{C}^{1}(\Omega)^{d_i}$.
\end{theorem} 
\begin{proof}
Since $\left(A,B\right) \in \mathcal{P}_{lti}^{state}$, $A$ is a block-diagonal matrix with say $L$ blocks.
We can thus partition the matrices $A$ and $B$ as
\begin{equation*}
    A = \textnormal{blkdiag}[A^{(1)}, \hdots, A^{(L)}], \quad B=\begin{bmatrix}
        [B^{(1)}]^T& 
        \hdots&
         [B^{(L)}]^T
    \end{bmatrix}^T,
\end{equation*}
where $A^{(l)} \in \mathbb{R}^{d_l \times d_l}$ and $B^{(l)} \in \mathbb{R}^{d_l \times d_i}$. 
Note that $A^{(l)}\in \mathcal{G}$ for all $l\in\{1, \cdots, L\}$ and $A^{(l)}$ has $k_r^{(l)}$ real eigenvalues and $k_c^{(l)}$ pairs of complex eigenvalues (with non-zero imaginary parts), where $d_l = k_r^{(l)} + 2 k_c^{(l)}$. Let $n_l:= k_r^{(l)} + k_c^{(l)}$ be the number of neurons in the horizontal layer $l$. 
Consider an arbitrary input $u \in \mathcal{C}^1(\Omega)^{d_i}$ to the LTI system producing the state $x \in \mathcal{C}^2(\Omega)^{d_h}$, i.e., $x=f_{lti}^s(u)$.
Partitioning the state as per the dimension of the blocks of $A$, let $x=\begin{bmatrix}
    [x^{(1)}]^T&\cdots&[x^{(L)}]^T
\end{bmatrix}^T$, where $x^{(l)}(t) \in \mathbb{R}^{d_l \times 1}$.
For a non-negative integer $a$, 
let $T^{(l)}_a = \begin{bmatrix}
			I_{a} \otimes \begin{bmatrix}
				1&0
			\end{bmatrix}\\
			I_{a} \otimes \begin{bmatrix}
				0&1
			\end{bmatrix}
		\end{bmatrix}$ and 
  $P^{(l)}=\begin{bmatrix}
			I_{k_r^{(l)}} & 0\\ 0 & T^{(l)}_{k_c^{(l)}}
		\end{bmatrix}$.
Lemma \eqref{lemma:m_lti_to_second_order} allows us to construct matrices $(M^{(l)}, C^{(l)}, K^{(l)}, E^{(l)}, V^{(l)})$ with the map $\mathfrak{m}_{lti}:(A^{(l)},B^{(l)}) \mapsto (M^{(l)}, C^{(l)}, K^{(l)}, E^{(l)}, V^{(l)})$ and 
matrices $(W^{(l)}, Q^{(l)}, Z^{(l)})$ with the map $\mathfrak{m}_{\eta} : (A^{(l)}, B^{(l)}) \mapsto (W^{(l)}, Q^{(l)}, Z^{(l)})$ such that the new variables $\xi^{(l)}(t)\in\mathbb{R}^{k_r+k_c}$, $\eta^{(l)}(t)\in\mathbb{R}^{k_c}$, $\xi_r^{(l)}(t)\in\mathbb{R}^{k_r}$ and $\xi^{(l)}_c(t)\in\mathbb{R}^{k_c}$ defined as 
\begin{align*}
		\left[\begin{array}{c c}
			[\xi^{(l)}(t)]^T &
			[\eta^{(l)}(t)]^T
		\end{array}\right]^T=\left[\begin{array}{c c c}
		[\xi_r^{(l)}(t)]^T &
		[\xi_c^{(l)}(t)]^T &
		[\eta^{(l)}(t)]^T
		\end{array}\right]^T=P^{(l)}x^{(l)}(t)
\end{align*}
	satisfy  $\xi^{(l)}(0)=0$, $\eta^{(l)}(0)=0$,
	\begin{align}
		&M^{(l)}\Ddot{\xi}^{(l)}(t)+C^{(l)}\dot{\xi}^{(l)}(t)+K^{(l)}\xi^{(l)}(t)=E^{(l)}u(t)+V^{(l)}\dot{u}(t),\\
		&\eta^{(l)}(t)=W^{(l)} \, \xi^{(l)}_c(t) + Q^{(l)} \dot{\xi}^{(l)}_c(t) + Z^{(l)} u(t),
	\end{align}
	for all $t \in \Omega$.
Now applying the converse statement of \cref{lemma: overall_dynamics_DyNN}, we can construct a DyNN with parameters $\left(\mathcal{M}^{(l)},\mathcal{C}^{(l)},\mathcal{K}^{(l)},\mathcal{W}^{(l)}\right)$ computed as $\mathfrak{n}_{dynn}^{-1}:(M^{(l)}, C^{(l)}, K^{(l)}, \newline E^{(l)}, V^{(l)}) \mapsto \left(\mathcal{M}^{(l)},\mathcal{C}^{(l)},\mathcal{K}^{(l)},\mathcal{W}^{(l)}\right)$ such that $\xi^{(l)}=f_{dynn}^{(l)}(u)$, $\dot{\xi}^{(l)}=\frac{d}{dt}\left(f_{dynn}^{(l)}(u)\right)$.
Thus, the map $\mathfrak{m}_h$ can be constructed as the composition $\mathfrak{n}_{dynn}^{-1} \circ \mathfrak{m}_{lti}$.

We now construct the parameters $\left(\Phi,\Psi\right) \in \mathcal{P}_{dynn}^{output}$ such that the output of the DyNN equals the output of the given LTI system. 
Since $P^{(l)}$ is a permutation matrix, we can reconstruct the state $x^{(l)}$ as
\begin{align*}
    x^{(l)}(t)&=\underbrace{\left[\begin{array}{cc|c}
			I_{k_r^{(l)}} & 0 & 0\\
			0 &	I_{k_c^{(l)}} \otimes \begin{bmatrix} 1&0 \end{bmatrix}^T&I_{k_c^{(l)}} \otimes \begin{bmatrix} 0&1 \end{bmatrix}^T
		\end{array}\right]}_{\begin{bmatrix}
		    P_{\xi}^{(l)}&\vline &P_{\eta}^{(l)}
		\end{bmatrix}}\left[\begin{array}{c}
			\xi^{(l)}(t)\\
			\hline
			\eta^{(l)}(t)
		\end{array}
		\right]\\
        &=P_{\xi}^{(l)}\xi^{(l)}(t)+P_{\eta}^{(l)}\eta^{(l)}(t)\\
        &=P_{\xi}^{(l)}\xi^{(l)}(t)+P_{\eta}^{(l)}\left(W^{(l)} \, \xi^{(l)}_c(t) + Q^{(l)} \dot{\xi}^{(l)}_c(t) + Z^{(l)} u(t)\right)\\
        &=P_{\xi}^{(l)}\xi^{(l)}(t)+P_{\eta}^{(l)}\left(\begin{bmatrix}
            0&W^{(l)}
        \end{bmatrix} \, \xi^{(l)}(t) + \begin{bmatrix}
            0&Q^{(l)}
        \end{bmatrix}\dot{\xi}^{(l)}(t) + Z^{(l)} u(t)\right)\\
        &=\begin{bmatrix}
            \left( P_{\xi}^{(l)} +  P_{\eta}^{(l)}\begin{bmatrix}
            0&W^{(l)}
        \end{bmatrix}\right) & \left(P_{\eta}^{(l)}\begin{bmatrix}
            0&Q^{(l)}
        \end{bmatrix}\right)
        \end{bmatrix}\begin{bmatrix}
            \xi^{(l)}(t) \\ \dot{\xi}^{(l)}(t)
        \end{bmatrix} + P_{\eta}^{(l)}Z^{(l)} u(t)\\
        &=\underbrace{\begin{bmatrix}
            \left( P_{\xi}^{(l)} +  P_{\eta}^{(l)}\begin{bmatrix}
            0&W^{(l)}
        \end{bmatrix}\right) & \left(P_{\eta}^{(l)}\begin{bmatrix}
            0&Q^{(l)}
        \end{bmatrix}\right)
        \end{bmatrix} T^{(l)}_{n_l}}_{\mathcal{F}^{(l)}}y^{(l)}(t) + P_{\eta}^{(l)}Z^{(l)} u(t),
\end{align*}
where, $y^{(l)}(t) = \begin{bmatrix}
            [y_1^{(l)}(t)]^T & \hdots & [y_{n_l}^{(l)}(t)]^T
        \end{bmatrix}^T $. 
Note that $\mathcal{F}^{(l)} \in \mathbb{R}^{d_l \times 2n_l}$, $T^{(l)}_{(n_l)} \in \mathbb{R}^{2n_l \times 2n_l}$, $P_{\eta}^{(l)} \in \mathbb{R}^{d_l \times k_c}$, and $Z^{(l)} \in \mathbb{R}^{k_c \times d_i}$. 
Stacking all $x^{(l)}$ to form the final state vector $x$ and plugging it into the output equation of the LTI system 
\begin{align*}
    y(t)&=Cx(t)+Du(t)\\
    &=C\left( \underbrace{\begin{bmatrix}
        \mathcal{F}^{(1)}&&\\
        &\ddots&\\
        &&\mathcal{F}^{(L)}
    \end{bmatrix}}_{\mathcal{F}}\begin{bmatrix}
            y^{(1)}(t)\\ \vdots \\ y^{(L)}(t)
        \end{bmatrix}+\underbrace{\begin{bmatrix}
        P_{\eta}^{(1)}Z^{(1)}(t)\\ \vdots \\ P_{\eta}^{(L)}Z^{(L)}(t)
    \end{bmatrix}}_{\mathcal{Z}}u(t) \right)+Du(t)\\
    &=C\mathcal{F}\begin{bmatrix}
            [y^{(1)}(t)]^T& \hdots & [y^{(L)}(t)]^T
        \end{bmatrix}^T+\left(C\mathcal{Z}+D\right)u(t).
\end{align*}

Comparing this with the output equation of the DyNN \eqref{eq:f_dynn} and matching coefficients, we obtain $\Psi=C\mathcal{Z}+D$ and
\begin{align}
    \left[\begin{array}{c c c|c c c|c|c c c}
            \phi_1^{(1)}&\cdots&\phi_{n_1}^{(1)}&\phi_1^{(2)}&\cdots&\phi_{n_2}^{(2)}&\cdots&\phi_1^{(L)}&\cdots&\phi_{n_L}^{(L)}
        \end{array}\right] = C\mathcal{F}
\end{align}
This completes the construction of the map $\mathfrak{m}_o$ and the proof.
\end{proof}
\begin{table}[ht]
\caption{Types of horizontal layers based on the number of real and complex eigenvalues $k_r^{(l)}, k_c^{(l)}$ of $A_{ll}$ (diagonal block $l$ of the transformed state matrix).  
Since each complex eigenvalue always appears in tandem with its conjugate, we represent the corresponding second-order neurons in a DyNN by yin-yang balls.
Only a part of the architecture corresponding to the term $A_{ll} \xi^{(l)}(t)$, which reveals connections within each hidden layer, is shown here (see \cref{fig:dynn_revised} for full architecture). The dashed self-connections indicate internal state dynamics.
}
\centering
\begin{tblr}
{colspec = {|Q[halign=c,valign=t,0.63\textwidth]|Q[halign=c,valign=t,0.34\textwidth]|}, 
width = \linewidth, 
colsep = 0pt}
    \hline
    State equation of the LTI system & Horizontal layer of DyNN \\ \hline
    \begin{minipage}[c]{0.59\textwidth}
    \begin{equation*}        
    \begin{gathered}
            \begin{bmatrix}
                \textcolor{green}{\dot{\xi}_{1}(t)} \\
                \textcolor{blue}{\dot{\xi}_{2}(t)} \\
                \textcolor{orange}{\dot{\xi}_{3}(t)} \\
                \textcolor{red}{\dot{\xi}_{4}(t)} \\
            \end{bmatrix}
            = 
            \underbrace{
            \begin{bmatrix}
                    \textcolor{green}{*} & \textcolor{green}{*} & \textcolor{green}{*} & \textcolor{green}{* }\\
                    & \textcolor{blue}{*} & \textcolor{blue}{*} & \textcolor{blue}{*} \\
                    & & \textcolor{orange}{*} & \textcolor{orange}{*} \\
                    & & & \textcolor{red}{*}\\
            \end{bmatrix}}_{A_{ll} \in \mathcal{G}_r}
            \begin{bmatrix}
                \textcolor{green}{{\xi}_{1}(t)} \\
                \textcolor{blue}{{\xi}_{2}(t)} \\
                \textcolor{orange}{{\xi}_{3}(t)} \\
                \textcolor{red}{{\xi}_{4}(t)} \\
            \end{bmatrix}
        + 
            B^{(l)}
            u(t).
        \end{gathered}
    \end{equation*}
    \end{minipage}
    & 
    \begin{minipage}[c]{0.3\textwidth}
                \centering
                \renewcommand{\thefigure}{3}
                \includegraphics[width=\textwidth]{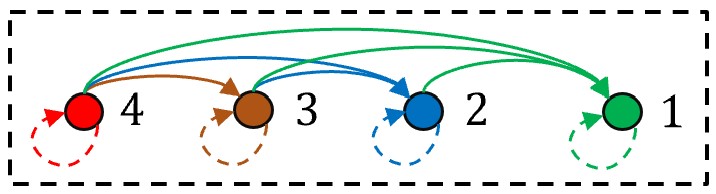}
                \captionof{figure}{$\mathbf{k_r^{(l)} = 4}, $ $\mathbf{k_c^{(l)} = 0}$: DyNN horizontal layer with four first-order neurons (solid balls).}
                \label{fig:real_k}
    \end{minipage}\\
        \hline
    \begin{minipage}[c]{0.59\textwidth}
    \begin{equation*}
    \begin{gathered}
    \begin{bmatrix}
                \textcolor{green}{\dot{\xi}_{1}(t)} \\
                \textcolor{green}{\dot{\xi}_{2}(t)} \\
                \textcolor{blue}{\dot{\xi}_{3}(t)} \\
                \textcolor{blue}{\dot{\xi}_{4}(t)} \\
                \textcolor{orange}{\dot{\xi}_{5}(t)} \\
                \textcolor{orange}{\dot{\xi}_{6}(t)} \\
                \textcolor{red}{\dot{\xi}_{7}(t)} \\
                \textcolor{red}{\dot{\xi}_{8}(t)} \\
            \end{bmatrix}
            = 
    \underbrace{\left[
        \begin{array}{c c c c c c c c} 
             \textcolor{green}{\ast} & \textcolor{green}{\ast} & \textcolor{green}{\ast} & \textcolor{green}{\ast} & \textcolor{green}{*} & \textcolor{green}{*} & \textcolor{green}{\ast} & \textcolor{green}{\ast}\\ 
             \textcolor{green}{\ast} & \textcolor{green}{\ast} & \textcolor{green}{\ast} & \textcolor{green}{\ast} & \textcolor{green}{*} & \textcolor{green}{*} & \textcolor{green}{\ast} & \textcolor{green}{\ast}\\ 
              &  & \textcolor{blue}{\ast} & \textcolor{blue}{\ast} & \textcolor{blue}{*} & \textcolor{blue}{*} & \textcolor{blue}{\ast} & \textcolor{blue}{\ast}\\ 
              &  & \textcolor{blue}{\ast} & \textcolor{blue}{\ast} & \textcolor{blue}{*} & \textcolor{blue}{*} & \textcolor{blue}{\ast} & \textcolor{blue}{\ast}\\ 
              &  &  &  & \textcolor{orange}{*} & \textcolor{orange}{*} & \textcolor{orange}{\ast} & \textcolor{orange}{\ast}\\ 
              &  &  &  &  \textcolor{orange}{*} & \textcolor{orange}{*} & \textcolor{orange}{\ast} & \textcolor{orange}{\ast}\\ 
             &  &  &  &  &  & \textcolor{red}{\ast} & \textcolor{red}{\ast}\\ 
              &  &  &  &  &  & \textcolor{red}{\ast} & \textcolor{red}{\ast}\\
          \end{array} 
        \right]}_{A_{ll} \in \mathcal{G}_c}
        \begin{bmatrix}
                \textcolor{green}{{\xi}_{1}(t)} \\
                \textcolor{green}{{\xi}_{2}(t)} \\
                \textcolor{blue}{{\xi}_{3}(t)} \\
                \textcolor{blue}{{\xi}_{4}(t)} \\
                \textcolor{orange}{{\xi}_{5}(t)} \\
                \textcolor{orange}{{\xi}_{6}(t)} \\
                \textcolor{red}{{\xi}_{7}(t)} \\
                \textcolor{red}{{\xi}_{8}(t)} 
        \end{bmatrix}
        \\+ 
            B^{(l)}
            u.
        \end{gathered}
        \end{equation*}
        \end{minipage}
        & 
        \begin{minipage}[c]{0.3\textwidth}
            \renewcommand{\thefigure}{4}
            \centering
            \includegraphics[width=\textwidth]{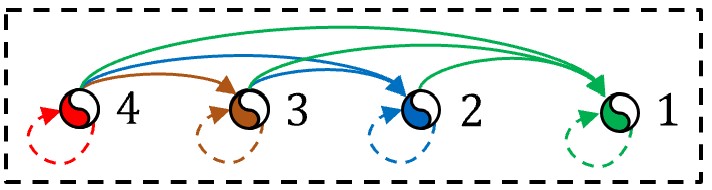}
        \captionof{figure}{$\mathbf{k_r^{(l)} = 0}, $ $\mathbf{k_c^{(l)} = 4}$: DyNN horizontal layer with four second-order neurons (yin-yang balls).}
        \label{fig:complex_k}
        \end{minipage}\\
        \hline
    \begin{minipage}[c]{0.59\textwidth}
    \begin{equation*}
    \begin{gathered}        
        \begin{bmatrix}
                \textcolor{green}{\dot{\xi}_{1}(t)} \\
                \textcolor{blue}{\dot{\xi}_{2}(t)} \\
                \textcolor{orange}{\dot{\xi}_{3}(t)} \\
                \textcolor{orange}{\dot{\xi}_{4}(t)} \\
                \textcolor{red}{\dot{\xi}_{5}(t)} \\
                \textcolor{red}{\dot{\xi}_{6}(t)} \\
        \end{bmatrix}
        =
        \underbrace{
        \left[ 
            \begin{array}{c c c c c c} 
                \textcolor{green}{\ast} & \textcolor{green}{\ast} & \textcolor{green}{\ast} & \textcolor{green}{\ast} & \textcolor{green}{*} & \textcolor{green}{*}\\ 
                 & \textcolor{blue}{\ast} & \textcolor{blue}{\ast} & \textcolor{blue}{*} & \textcolor{blue}{*} & \textcolor{blue}{\ast}\\ 
                  &  & \textcolor{orange}{\ast} & \textcolor{orange}{\ast} & \textcolor{orange}{*} & \textcolor{orange}{*}\\ 
                  &  & \textcolor{orange}{\ast} & \textcolor{orange}{\ast} & \textcolor{orange}{*} & \textcolor{orange}{*}\\ 
                 &  &  &  &  \textcolor{red}{\ast} & \textcolor{red}{\ast}\\ 
                 &  &  &  &  \textcolor{red}{\ast} & \textcolor{red}{\ast}\\
              \end{array} 
            \right]}_{A_{ll} \in \mathcal{G}}
            \begin{bmatrix}
                \textcolor{green}{{\xi}_{1}(t)} \\
                \textcolor{blue}{{\xi}_{2}(t)} \\
                \textcolor{orange}{{\xi}_{3}(t)} \\
                \textcolor{orange}{{\xi}_{4}(t)} \\
                \textcolor{red}{{\xi}_{5}(t)} \\
                \textcolor{red}{{\xi}_{6}(t)} 
        \end{bmatrix}
        \\+ 
            B^{(l)}
            u(t).
        \end{gathered}
        \end{equation*}
        \end{minipage}
        & 
        \begin{minipage}[c]{0.3\textwidth}
        \centering
        \renewcommand{\thefigure}{5}
        \includegraphics[width=\textwidth]{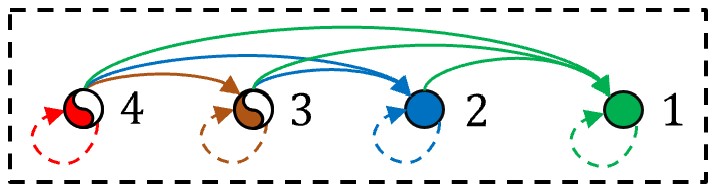}
        \captionof{figure}{$\mathbf{k_r^{(l)} = 2}, $ $\mathbf{k_c^{(l)} = 2}$: DyNN horizontal layer with two first-order neurons (green and blue solid balls) and two second-order neurons (red and dark-orange yin-yang balls).}
        \label{fig:real_complex_k}
        \end{minipage}\\
    \hline
\end{tblr}
\label{Tab:types_horizontal_layers}
\vspace{-2em}
\end{table}
\addtocounter{figure}{-9} 
\cref{Tab:types_horizontal_layers} illustrates how the sparsity patterns of the diagonal blocks (see \cref{def:sparse_G}) of the transformed state matrix $A$ result in different types of horizontal hidden layers of the dynamic neural network. 
Note that in the second row of \cref{Tab:types_horizontal_layers}, we map each pair of first-order ODEs shown in the same color (corresponding to a pair of complex eigenvalues of $A$) to a second-order ODE (see \cref{lemma:m_lti_to_second_order}) and represent it by a corresponding second-order neuron (see  \cref{th:parameter_maps}). 
The proposed mapping requires us to differentiate one of the state equations with respect to time, leading to terms involving $\dot{u}$ (see \cref{lemma:m_lti_to_second_order}).

%% file: sections/algorithm.tex
\subsection{Dynamic neural network algorithm} \label{ss:algorithm}
We present two algorithms that summarize our implementation of dynamic neural networks in this section.  
\cref{alg:construct_dynamic_neural_network} accepts a state-space model ($\tilde{A}, \tilde{B}, \tilde{C}, \tilde{D}$) as input and, for a selected \texttt{clustering algorithm}, constructs a dynamic neural network architecture together with its parameters. 
The input is pre-processed as described in \cref{alg:preprocessing_lti}. 
Based on the number of real and complex eigenvalues within each diagonal block of the transformed state matrix $A$, we construct an appropriate horizontal layer as shown in \cref{Tab:types_horizontal_layers}. 
Finally, all parameters of horizontal layers and the output layer of the DyNN are computed using the maps $\mathfrak{m}_h$ and $\mathfrak{m}_o$ as described in \cref{sec:all_mappings}. 

\cref{alg:dynamic neural network} accepts as input a DyNN with fixed architecture and parameters (output of \cref{alg:construct_dynamic_neural_network}), inputs $u(t), \dot{u}(t)$, and time domain $\Omega = [t_0, t_f]$ with initial and final times $t_0$, $t_f$ respectively, and describes how to compute the output of a DyNN. 
If $u$ is available only at a finite number of time points, then the user can specify how to interpolate. 
Currently, we provide an option to interpolate $u$ with either a piecewise constant function or a piecewise linear function.
The initial conditions of the ODE to be solved for the state of each neuron are set to zero. 

The user can choose the ODE solver denoted by the parameter \texttt{method} from any of the standard explicit solvers, e.g., RK45 \cite{dormand1980family}, RK23 \cite{shampine1986some}, DOPRI85 \cite[section 2.5]{hairer1993solving} or implicit solvers, e.g, Radau \cite{hairer-1996}, BDF \cite{shampine1997matlab}, LSODA \cite{petzold1983automatic}, and many more that are implemented in the \texttt{solve\_ivp} routine of the SciPy package \cite{2020SciPy-NMeth}.
The parameter \texttt{dense\_output} of the method \texttt{solve\_ivp} is set to true, which means that the output of the ODE is a function handle that can be evaluated by interpolation at any time point $t \in \Omega$. 
The order of interpolation depends on the method specified. 
For instance, for RK23, a cubic Hermite polynomial is used. 
For DOPRI85, a seventh-order polynomial is used.  
The user can also specify relative and absolute tolerances for the solver denoted by \texttt{rtol}, \texttt{atol}. 
Note that parameters \texttt{method}, \texttt{rtol}, \texttt{atol} can even be different for different neurons. 
The output of the DyNN is then computed as a function handle.
Note that line 10 in the \cref{alg:dynamic neural network} concerning the output $\hat{y}$ is a functional assignment. 
The user can specify the time points at which the output will be evaluated. 
See \cref{app_rem:comp_time_derivatives} for details on computing the time-derivative of the input and \cref{app:eff_imple_firstorder} for efficient implementation of first-order neurons.

\newcommand{\COMMENT}[2][.2\linewidth]{%
  \leavevmode\hfill\makebox[#1][l]{//~#2}}
\begin{algorithm}[ht]
    \caption{Computing dynamic neural network architecture and parameters}\label{alg:construct_dynamic_neural_network}
    \textbf{Input:}  State Space Model ($\tilde{A}, \tilde{B}, \tilde{C}, \tilde{D}$)\\
    \textbf{Output:} DyNN parameters - $\left(\mathcal{M},\mathcal{C},\mathcal{K},\mathcal{W},\Theta\right) \in \mathcal{P}_{dynn}^{hidden}, \left(\Phi,\Psi\right) \in \mathcal{P}_{dynn}^{output}$\\
    \textbf{Parameters:} \texttt{clustering algorithm}
    \begin{algorithmic}[1]
    \State \textbf{Pre-process the LTI system}
    \State $\quad$ Pre-process the LTI system:
    ($A, B, C, D$) $ \, \gets $ ($\tilde{A}, \tilde{B}, \tilde{C}, \tilde{D}$)   \COMMENT{\cref{alg:preprocessing_lti}}
    \State \textbf{Construct horizontal layers of DyNN}
        \For{$l \gets 1$ to $L$}
            \State Construct a layer with $k_r^{(l)}$, $k_c^{(l)}$ first- and second-order neurons \COMMENT{\cref{Tab:types_horizontal_layers}} 
            \State Compute parameters  $\left(\mathcal{M}^{(l)},\mathcal{C}^{(l)},\mathcal{K}^{(l)},\mathcal{W}^{(l)}\right)$  $\in \mathcal{P}_{dynn}^{(l)}$  \COMMENT{\cref{th:parameter_maps}}
        \EndFor
    \State \textbf{Construct output layer of DyNN}
        \State Compute output layer parameters $\left(\Phi,\Psi\right) \in \mathcal{P}_{dynn}^{output}$ \COMMENT{\cref{th:parameter_maps}}
    \end{algorithmic}
\end{algorithm}  
\vspace{-1em}
\renewcommand{\COMMENT}[2][.17\linewidth]{%
  \leavevmode\hfill\makebox[#1][l]{//~#2}}
    \begin{algorithm}[ht]
    \caption{Forward pass of a dynamic neural network}\label{alg:dynamic neural network}
    \textbf{Input:} DyNN architecture and parameters - $\left(\mathcal{M},\mathcal{C},\mathcal{K},\mathcal{W},\Theta\right) \in \mathcal{P}_{dynn}^{hidden}, \left(\Phi,\Psi\right) \in \mathcal{P}_{dynn}^{output}$, inputs $u$ and $\dot{u}$ as function handles, time domain $\Omega = [t_0, t_f]$\\
    \textbf{Output:} Output of the dynamic neural network $\hat{y}$ as a function handle\\
    \textbf{Parameters:} \texttt{method, rtol, atol}
    \begin{algorithmic}[1]
        \For{$l \gets 1$ to $L$}
            \For{$i \gets n_l$ to $1$} 
                \State Set initial conditions $\hat{y}_i^{(l)}(0)$ to $0$. 
                \State \texttt{properties} $\gets$ \texttt{method, rtol, atol}
                \State \texttt{weights} $\gets \left( m^{(l)}_i,c^{(l)}_i,k^{(l)}_i, w_i^{(l)}, \phi_i^{(l)}
                    \right)$
                \State $\hat{u}_i^{(l)}$ $\gets$ $\begin{bmatrix} u^T&\dot{u}^T&(\hat{y}_{i+1}^{(l)})^T&(\hat{y}_{i+2}^{(l)})^T&\cdots&(\hat{y}_{n_l}^{(l)})^T
                \end{bmatrix}^T  $ 
                \State $\hat{y}_i^{(l)}$ $\gets$ \texttt{solve\_ivp}($\hat{y}_i^{(l)}(0), \hat{u}_i^{(l)}$, $\Omega$, 
                \texttt{weights}, \texttt{properties})     
            \EndFor
        \EndFor
    \State Compute DyNN output $\hat{y}\gets\biggl(\sum_{l=1}^{L} \sum_{i=1}^{n_l} \phi_i^{(l)}  \, \hat{y}_i^{(l)} \biggr) + \Psi \, u$ 
    \end{algorithmic}
\end{algorithm}  

In analogy with input-output maps defined in \cref{ss:dynn} based on analytical solutions of the ODEs, we now define the input-output maps for a neuron and a DyNN based on numerical solutions of the ODEs. 
These maps are then used for the error analysis presented in the next subsection.
\begin{defn}[Input-output map of a numerically implemented neuron] \label{def_io_neuron_num}
The \textbf{input-output map of the numerically implemented neuron} $i$ in hidden layer $l$ with $d_i^{(l)}$ inputs is a map $\hat{f}_i^{(l)}:\mathcal{C}(\Omega)^{d_i^{(l)}}  \rightarrow \mathcal{C}^1(\Omega)^{2}$ defined as $\hat{u}_i^{(l)}\mapsto \hat{y}_i^{(l)}$, where $\hat{y}_i^{(l)}$ is the output of the function \textnormal{\texttt{solve\_ivp}} used in \cref{alg:dynamic neural network} corresponding to input $\hat{u}_i^{(l)}$ and parameters $(m^{(l)}_i,c^{(l)}_i,k^{(l)}_i, w_i^{(l)})$.
\end{defn}
\begin{defn}[Input-output map of a numerically implemented DyNN]\label{def_io_dynn_num}
    Corresponding to a given DyNN and parameters of \cref{alg:dynamic neural network}, the \textbf{input-output map of a numerically implemented DyNN} with $L$ horizontal layers, $n_l$ neurons in the horizontal layer $l$, $d_i$ neurons in the input layer and $d_o$ neurons in the output layer is defined as $\hat{f}_{dynn}:u\mapsto \hat{y}$ where $\hat{y}$ is the output of \cref{alg:dynamic neural network} corresponding to parameters $\left(\mathcal{M},\mathcal{C},\mathcal{K},\mathcal{W},\Theta\right) \in \mathcal{P}_{dynn}^{hidden}, \left(\Phi,\Psi\right) \in \mathcal{P}_{dynn}^{output}$ and inputs $u$ and $\dot{u}$.
\end{defn}

%% file: sections/numerical_analysis.tex
\subsection{Error analysis for dynamic neural networks}\label{ss:error_analysis}
We now provide an upper bound on the numerical error of the DyNN compared to the analytical solution. 
We assume that the continuous extensions of ODE solvers used in \texttt{solve\_ivp} (by setting \texttt{dense\_output} to true) satisfy certain error bounds (see \eqref{eq:assumption_solve_ivp}). 
For continuous extensions of ODE solvers like Runge-Kutta methods and Dormand-Prince methods, which approximate the solution at any point in the span of a time step and the corresponding error bounds, we refer the reader to \cite{horn1983fourth, owren1991order} and \cite[Chapter 2]{hairer1993solving}.
\begin{figure}[h]
    \begin{center}
        \begin{minipage}{\textwidth}
        \centering
            \includegraphics[width=0.87\textwidth]{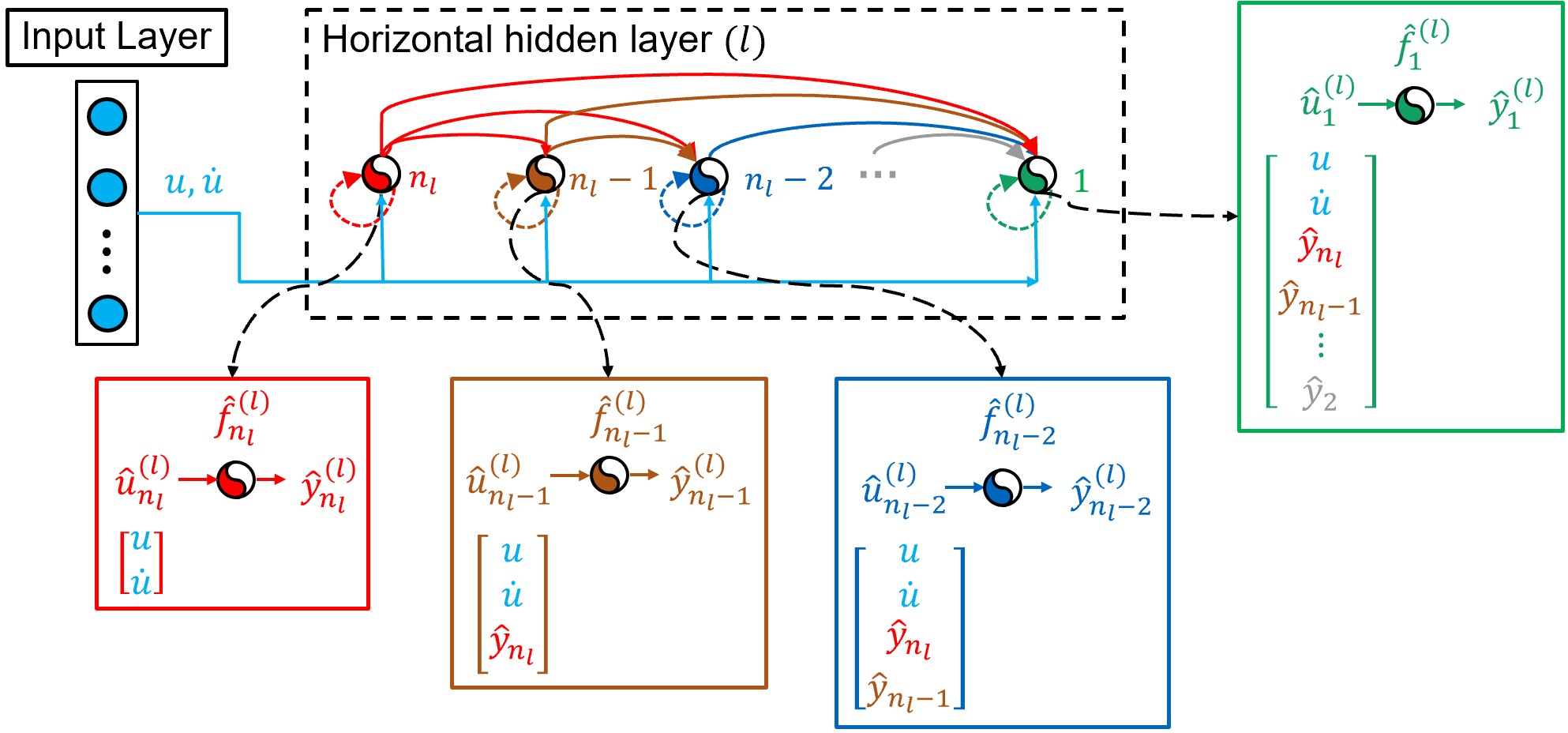}
            \caption{Illustration of the input-output maps of all neurons in a dynamic neural network's horizontal hidden layer $l$. The output layer connections are omitted for brevity.}
            \label{fig:dynn_connections_io} 
        \end{minipage}
    \end{center}
\vspace{-1.7em}
\end{figure}
\begin{theorem}
\label{th:numerical_analysis}
    Assume that function \textnormal{\texttt{solve\_ivp}} (line $7$ of \cref{alg:dynamic neural network}) implemented on all neurons $i\in \{1,\cdots,n_l\}$ in all horizontal hidden layers $l\in\{1,\cdots,L\}$ mapping the input $\hat{u}_i^{(l)}$ to the solution $\hat{y}_i^{(l)}$ satisfies
    \begin{equation}\label{eq:assumption_solve_ivp}
         || \hat{y}_i^{(l)}(t) - f_i^{(l)}(\hat{u}_i^{(l)})(t)|| = \mathcal{O}(h^p) \quad \textnormal{as} \,\, h \to 0, \, \forall t \in \Omega, \, \forall \hat{u}_i^{(l)} \in \mathcal{C}^1(\Omega)^{d_i^{(l)}},
    \end{equation}
    where $f_i^{(l)}$ is the input-output map corresponding to neuron $i$ in layer $l$ (see \cref{def_io_neuron}). 
    Then we have that
    \begin{equation*}
        ||f_{dynn}(u)(t) - \hat{f}_{dynn}(u)(t) || = \mathcal{O}(h^p) \quad \textnormal{as} \,\, h \to 0, \, \forall t \in \Omega, \, \forall u \in \mathcal{C}^1(\Omega)^{d_i},
    \end{equation*}
    where $f_{dynn}$ is the input-output map of the dynamic neural network (see \cref{def_iso_dynn}) and $\hat{f}_{dynn}$ is the input-output map of the numerical implementation of the dynamic neural network (see \cref{def_io_dynn_num}), both corresponding to the same parameters.
\end{theorem}
\begin{proof}
For a neuron $i$ in horizontal layer $l$ of a DyNN, consider an arbitrary input $u_i^{(l)}(t)$ corrupted by noise (error) $\tilde{u}_i^{(l)}(t)$ and define $\hat{u}_i^{(l)}(t) = u_i^{(l)}(t) + \tilde{u}_i^{(l)}(t)$. 
First, we bound $\fnorm{f_i^{(l)}(u_i^{(l)}) - f_i^{(l)}(\hat{u}_i^{(l)})}$, which is the error in the output of the map $f_i^{(l)}$ due to the noise in the input. 
Second, we bound $\fnorm{f_i^{(l)}(u^{(l)}_i) - \hat{f}_i^{(l)}(\hat{u}^{(l)}_i)}$, which takes into account the noise in the input for neuron $i$ of layer $l$, and the numerical error introduced by the map $\hat{f}_i^{(l)}$. 
Finally, we recursively bound the error accumulated by the successive neurons in a given horizontal layer.

Since the input-output map $f_i^{(l)}$ is defined via solution to a linear, time-invariant differential equation with zero initial conditions (see \cref{def_io_neuron}), it can be shown that there exists a $C_i^{(l)} \in \mathbb{R}$ such that 
\begin{align}\label{eq:neuron_bound}
        \fnorm{f_i^{(l)}(u_i^{(l)}) - f_i^{(l)}(\hat{u}_i^{(l)})}= \fnorm{f_i^{(l)}(\tilde{u}_i^{(l)})}\le C_i^{(l)} \fnorm{\tilde{u}_i^{(l)}}.
    \end{align}
Now observe that for any $l\in\{1,\cdots,L\}$,
\begin{align}
        \fnorm{f_i^{(l)}(u^{(l)}_i) - \hat{f}_i^{(l)}(\hat{u}^{(l)}_i)} &= \fnorm{f_i^{(l)}(u_i) - f_i^{(l)}(\hat{u}^{(l)}_i) + f_i^{(l)}(\hat{u}^{(l)}_i) -\hat{f}_i^{(l)}(\hat{u}^{(l)}_i)} \nonumber \\
        &\le \fnorm{f_i^{(l)}(u_i) - f_i^{(l)}(\hat{u}^{(l)}_i)} + \fnorm{f_i^{(l)}(\hat{u}^{(l)}_i) -\hat{f}_i^{(l)}(\hat{u}^{(l)}_i)} \nonumber\\
        & \le C_i^{(l)} \fnorm{\tilde{u}_i^{(l)}} + \fnorm{f_i^{(l)}(\hat{u}^{(l)}_i) -\hat{f}_i^{(l)}(\hat{u}^{(l)}_i)}, \nonumber 
\end{align}
For constants
$\varepsilon \coloneqq \max_{i,l}\fnorm{f_i^{(l)}(\hat{u}^{(l)}_i) -\hat{f}_i^{(l)}(\hat{u}^{(l)}_i)}$ and $C \coloneqq\max_{i,l} C_i^{(l)}$, we get 
\begin{align}
        \fnorm{f_i^{(l)}(u^{(l)}_i) - \hat{f}_i^{(l)}(\hat{u}^{(l)}_i)} 
        \label{eq:ub_across_neuron}
        & \le C \fnorm{\tilde{u}_i^{(l)}} + \varepsilon. 
\end{align}
Due to the topology of neurons in any horizontal layer of a DyNN as described by the input-output map of each neuron and depicted in \cref{fig:dynn_connections_io} for a single hidden layer, note that
\begin{align}\label{eq:error_free_input_bound}
    \fnorm{\tilde{u}_i^{(l)}}=\fnorm{u_i^{(l)}-\hat{u}_i^{(l)}}&=\max_{j\in\{i+1,\cdots, n_l\}}\fnorm{y_{j}^{(l)}-\hat{y}_{j}^{(l)}}\\
    &=\max_{j\in\{i+1,\cdots, n_l\}}\fnorm{f_j^{(l)}(u^{(l)}_j) - \hat{f}_j^{(l)}(\hat{u}^{(l)}_j)}\\
    &=\fnorm{f_{k_1}^{(l)}(u^{(l)}_{k_1}) - \hat{f}_{k_1}^{(l)}(\hat{u}^{(l)}_{k_1})}
\end{align}
for some $k_1\in\{i+1,\cdots, n_l\}$.
Using the bound \eqref{eq:ub_across_neuron} recursively, we obtain
\begin{align*}
    \fnorm{\tilde{u}_i^{(l)}}\le C^2\fnorm{\tilde{u}_{k_2}^{(l)}} + C\varepsilon+\varepsilon&\le C^3\fnorm{\tilde{u}_{k_3}^{(l)}} + C^2\varepsilon  + C\varepsilon+\varepsilon\\
    &\vdots\\
    &\le \left(1+C+C^2+\cdots+C^{k_m}\right)\varepsilon
\end{align*}
for some sequence $k_j$ with $i<k_1<k_2\cdots<k_m\leq n_l$.
Thus, there exists a constant $C_i$ such that
\begin{equation*}
    \fnorm{f_i^{(l)}(u^{(l)}_i) - \hat{f}_i^{(l)}(\hat{u}^{(l)}_i)}\leq C_i\varepsilon.
\end{equation*}
Using the definition of the input-output map of a DyNN (see \cref{def_iso_dynn}) and assumption \eqref{eq:assumption_solve_ivp}, there exists a constant $\Bar{C}$ such that 
\begin{equation*}
    \fnorm{f_{dynn}(u) - f_{dynn}(\hat{u})}\leq \Bar{C} \varepsilon=\Bar{C}\max_{i,l}\fnorm{f_i^{(l)}(\hat{u}^{(l)}_i) -\hat{f}_i^{(l)}(\hat{u}^{(l)}_i)}, 
\end{equation*}
which implies that $ ||f_{dynn}(u)(t) - \hat{f}_{dynn}(u)(t) || = \mathcal{O}(h^p) \,\,\, \textnormal{as} \,\, h \to 0, \, \forall t \in \Omega$.
\end{proof}
\begin{remark}
    We would like to emphasize that the proof does not rely on the linearity of the map $f_i^{(l)}$, but instead on the existence of a constant $C_i^{(l)}$ such that \eqref{eq:neuron_bound} holds.
    Similarly, the map of the output layer may be non-linear as long as it is Lipschitz continuous. 
    This gives us a natural way to extend the above analysis to non-linear systems in the future.
    Secondly, if the terms $u, \dot{u}$ are to be approximated, one can extend the analysis to include this additional source of error.    
\end{remark}

%% file: sections/numerical_results.tex
\section{Numerical Results} 
In this section, we showcase the accuracy of our dynamic neural networks in simulating LTI systems using diverse examples to illuminate various facets of our algorithm and validate our systematic approach to neural architecture construction.
We pre-process the LTI system with \cref{alg:preprocessing_lti}, construct a suitable dynamic neural network with \cref{alg:construct_dynamic_neural_network},  and perform forward pass with \cref{alg:dynamic neural network} to compute the output of our network. 
We intend to use these examples as proof of concept and emphasize that the goal here is not to outperform the existing solvers for simulating LTI systems.
The code and data for all the numerical examples, along with the details on parameter settings, are made available \footnote{URL for code and data: \url{https://gitlab.com/chinmay_datar/dynamic-neural-networks.git}}. 
\label{sec:numerical_results}
\subsection{Diffusion Equation: Unitarily diagonalizable state matrix}
\label{ss:diffusion}
A two-dimensional transient diffusion equation is 
\begin{align} \label{eq:diff}
    \frac{\partial T}{\partial t}(x, y, t) = \mathcal{D} \Bigl(\frac{\partial^2 T}{\partial x^2}(x, y, t) + \frac{\partial^2 T}{\partial y^2}(x, y, t)\Bigr) + \mathcal{S}(x, y, t), 
\end{align}
where T is the variable of interest (concentration of species or temperature), $\mathcal{D}$ is diffusivity, and $\mathcal{S}$ is the source term.
We interpret this as a system with $\mathcal{S}$ as the input and the solution $T$ as the output.
The boundary conditions are periodic, and the initial condition is $T(x,y,0) = 0$. 
Heat is injected into the system via the source term $\mathcal{S}(x, y, t)$ which is obtained by piecewise linear interpolation in time of the function $100 \exp{\Bigl( -0.8\bigl((x-l/2)^2 +(y-l/2)^2)\bigr)\Bigr)}\delta(t-0.2)$, where $\delta$ is the discrete-time unit impulse.
See \cref{app:ps_diff} for a detailed problem setup. 

We discretize equation \eqref{eq:diff} in space and get an LTI system of the form \cref{eq:lti_0}.  
We use a finite difference discretization on a uniform grid and obtain a symmetric state matrix. 
All symmetric matrices are unitarily diagonalizable and have real eigenvalues. Thus, the real Schur form in \cref{alg:preprocessing_lti} yields a transformed state matrix that is diagonal, and the dynamics across each state variable are decoupled.

\begin{figure}
    \begin{minipage}[t]{0.48\textwidth}
        \centering
        \vspace{0pt}
        \includegraphics[width=0.66\textwidth]{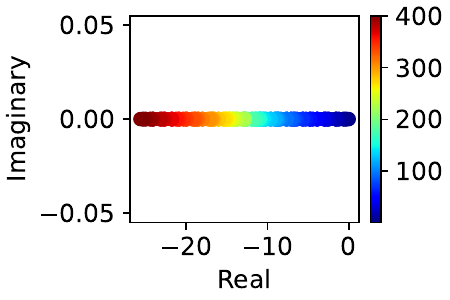}
        \captionof{figure}{Eigenvalues of the state matrix ordered as shown in the color bar (Ex 3.1).
        }
        \label{fig:eigenvalue_clustering_diff}
    \end{minipage}\hfill
    \begin{minipage}[t]{0.48\textwidth}
        \centering
        \vspace{0pt}
        \includegraphics[width=0.55\textwidth]{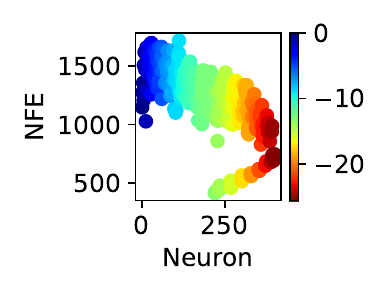}
        \captionof{figure}{Number of Function Evaluations (NFE) in ODEs for each neuron. Colors depict eigenvalues from \cref{fig:eigenvalue_clustering_diff} (Ex 3.1).
        }
        \label{fig:n_diff_nfe}
    \end{minipage}
\vspace{-1.8em}
\end{figure}

\begin{figure}[htbp]
\begin{minipage}[b]{0.32\textwidth}
        \centering
        \includegraphics[width=0.95\textwidth]{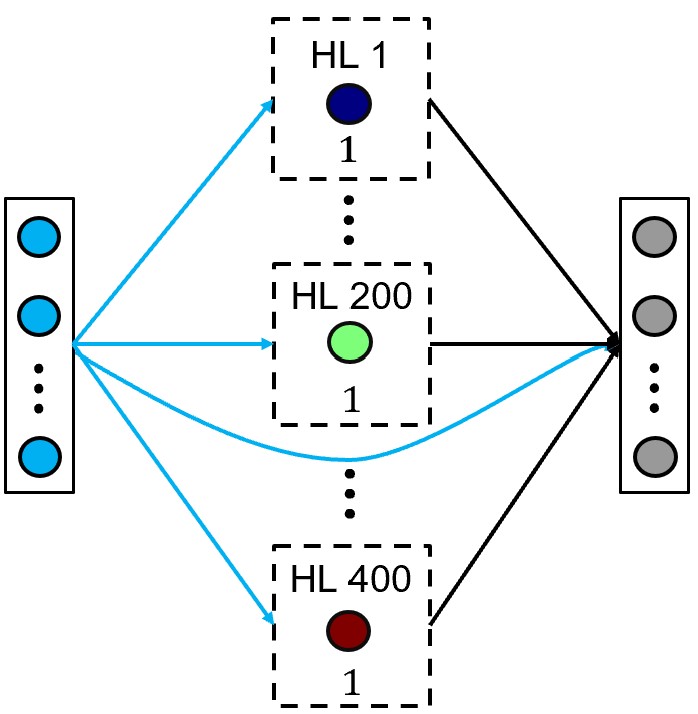}
        \captionof{figure}{DyNN architecture. Colors show different Horizontal Layers (HLs) (Ex 3.1).
        }
        \label{fig:diff_arch}
    \end{minipage}\hfill
    \begin{minipage}[b]{0.63\textwidth}
        \centering
        \includegraphics[width=0.85\textwidth]{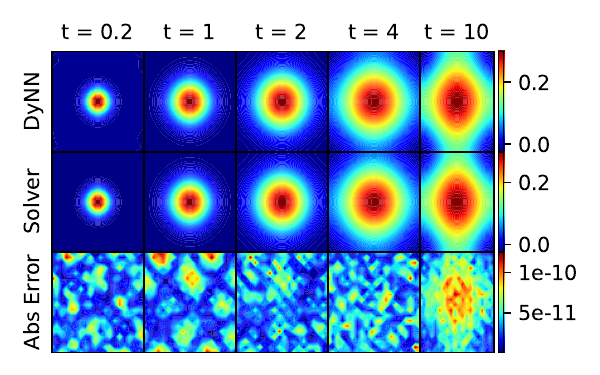}   
        \captionof{figure}{Top Panel: DyNN solution. Middle panel: numerical solution. Bottom panel: absolute error between the two solutions at five time instants (Ex 3.1).}
        \label{fig:n_diffusion_solution}
    \end{minipage}
\vspace{-1.8em}
\end{figure}

Since each diagonal block of the transformed state matrix is of size $1 \times 1$, \cref{alg:construct_dynamic_neural_network} constructs a DyNN architecture with each horizontal layer consisting of a single first-order neuron, i.e., an architecture with a single vertical layer as shown in \cref{fig:diff_arch}. 
As the transformed LTI system is mapped to the DyNN formalism, each first-order ODE is represented by a corresponding first-order neuron. 
\cref{fig:eigenvalue_clustering_diff} shows the eigenvalues ranging from -25.6 to 0 that naturally appear on the diagonal of the transformed state matrix sorted in ascending order according to the absolute value of the eigenvalue.  
During the forward pass of the DyNN, an ODE is solved for each neuron (line 7 of \cref{alg:dynamic neural network}). 
\cref{fig:n_diff_nfe} shows that the Number of Function Evaluations (NFE) or equivalently evaluations of the right-hand sides of the ODEs corresponding to different neurons vary a lot for different neurons. 
As we can use different ODE solvers for each neuron, this variability in the NFE can be exploited.
If the system is solved in its initial LTI form without decoupling, the same explicit ODE solver requires an NFE count of $1739$, each involving evaluation of the entire state matrix $A$. Whereas, in the DyNN, the maximum NFE count amongst all neurons is $1718$, each involving evaluation of only a $1 \times 1$ diagonal block of state matrix, substantially reducing the computational cost.
Finally, the comparison with the numerical simulation of the LTI system using a Python routine \texttt{SciPy.signal.lsim}  \cite{2020SciPy-NMeth} presented in \cref{fig:n_diffusion_solution} illustrates DyNN's ability to accurately simulate the semi-discretized diffusion equation.

\subsection{The reason for horizontal layers} \label{ss:horizontal_layers}
The goal of the next numerical example is to answer the following two questions: 
\begin{itemize}
    \item What happens if we choose a very high number of eigenvalue clusters in the clustering algorithm (or the number of horizontal layers in a DyNN) to aggressively enforce sparsity in neural connections?
    \item How does one choose a suitable number of clusters in the pre-processing \cref{alg:preprocessing_lti}? 
\end{itemize}

We consider a state matrix that is not unitarily diagonalizable with $k_r$ real eigenvalues and $k_c$ pairs of complex eigenvalues.  
For a given LTI system, ideally, we want to construct the sparsest possible dynamic neural network, i.e., a DyNN with a single vertical layer (all horizontal layers with exactly one neuron), which can represent the LTI system. 
This requires the matrix $\tilde{R}$ in the pre-processing \cref{alg:preprocessing_lti} to form $k_r + k_c$ eigenvalue clusters.
However, if the eigenvalues in different diagonal blocks of the matrix $\tilde{R}$ in \cref{alg:preprocessing_lti} 
are the same (see line 7 of \cref{alg:preprocessing_lti}), the assumption of the Bartels-Stewart algorithm is violated. 
Even if they are distinct but close to each other, the Bartels-Stewart algorithm yields a highly ill-conditioned transformation matrix $\mathcal{T}$. 
This explains why one must regroup close or identical eigenvalues together in respective clusters. 
Clustering trades off sparsity for a lower condition number of the transformation matrix $\mathcal{T}$ and results in the transformed state matrix
$A$ with fewer than $k_r + k_c$ diagonal blocks. 
We now validate the considerations of numerical stability and answer the outlined questions empirically. 

We construct an LTI system with input, state, and output dimensions $d_i = d_h = d_o = 10$. 
The state matrix $\Tilde{A} \in \mathbb{R}^{10 \times 10}$ is upper-triangular, with all entries above the diagonal sampled uniformly from $[0, 0.1]$. Importantly, the diagonal entries, which are also the eigenvalues, are chosen as  
$\lambda_n = -4 + \left(2.5\right)^{-n}$ for 
$i \in \{1, 2, \hdots, 10\}$. 
Please refer to \cref{app:ps_sp_sc} for the detailed problem setup. 

\begin{figure}[htbp]
\begin{centering}
\includegraphics[width=\textwidth]{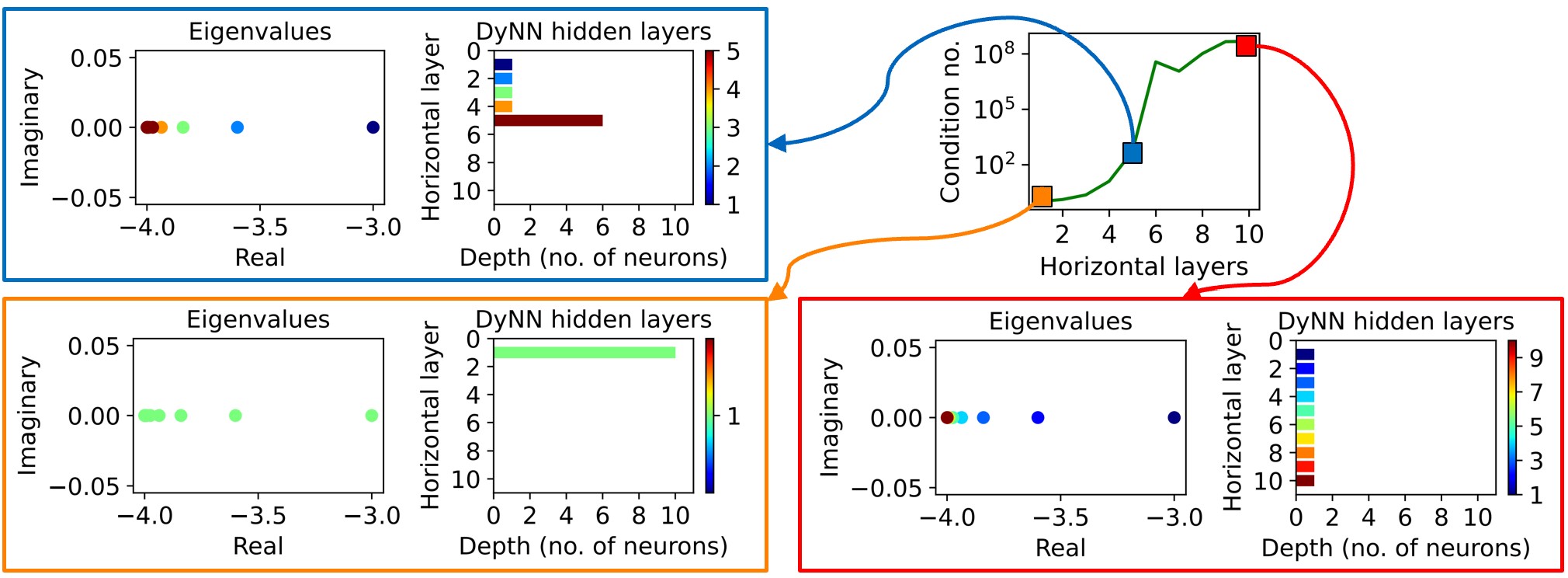}   
    \captionof{figure}{Top right: Growth of condition number ($C_t$) of the transformation matrix $\mathcal{T}$ with the number of horizontal layers in the constructed DyNN. 
    The orange, blue, and red boxes illustrate eigenvalue clustering with 1, 5, and 10 clusters (and the corresponding DyNN architectures with 1, 5, and 10 horizontal layers), respectively. 
    Within each box, horizontal layers and the corresponding eigenvalue clusters are marked with the same color (Ex 3.2).}
    \label{fig:cond_arch}
    \end{centering}
\vspace{-1.5em}
\end{figure}

\begin{figure}[h]
    \begin{center}
    \begin{minipage}[b]{0.3\textwidth}
            \centering
            \includegraphics[width=0.9\textwidth]{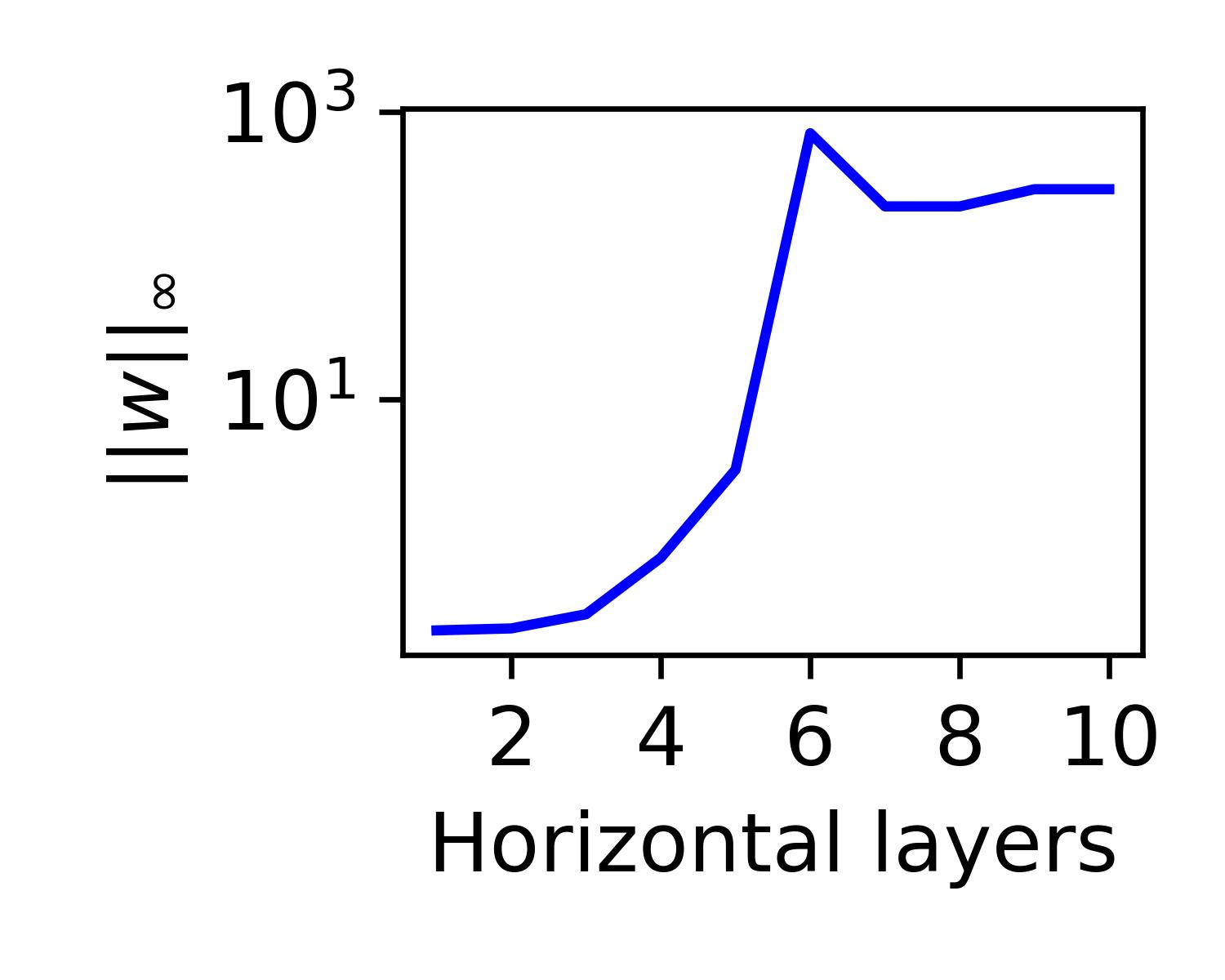}
            \caption{Blow-up of the network parameter values with horizontal layers (Ex 3.2).}
            \label{fig:weight_norm}
    \end{minipage}\hfill
    \begin{minipage}[b]{0.65\textwidth}
            \centering
            \includegraphics[width=\textwidth]{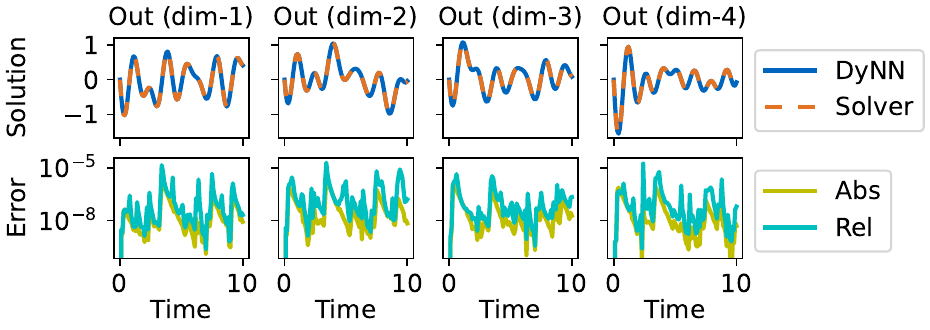}   
            \captionof{figure}{Top panel: Outputs of DyNN and numerical solver.
            Bottom panel: relative and absolute errors between the DyNN output and the numerical solution (Ex 3.2).}
            \label{fig:sol_sparsity_cond}
    \end{minipage}
    \end{center}
\vspace{-1.2em}
\end{figure}

We construct ten dynamic neural networks with different architectures corresponding to different numbers of eigenvalue clusters (number of horizontal layers) ranging from one to ten. 
\cref{fig:cond_arch} demonstrates how the condition number of the transformation matrix $\mathcal{T}$ from \cref{alg:preprocessing_lti} blows up as one increases the number of clusters of eigenvalues. 
Secondly, the high condition numbers of transformation matrices often lead to pre-processed matrices with very high values. 
\cref{fig:weight_norm} shows that the network weights blow up, too, as the number of horizontal layers increases.

In general, the smallest difference between eigenvalues of the state matrix that are in different clusters, along with the off-diagonal entries of the matrix $\mathcal{R}$, together affect the condition number of the transformation matrix.
We advocate for the implementation of a user-defined tolerance on the condition number of the transformation matrix as a means to determine a suitable number of clusters for the given LTI system. 
In this example, for instance, if the acceptable threshold on the condition number of the transformation matrix $C_t$ is $15$, one should select an architecture with 4 horizontal layers.
Finally, \cref{fig:sol_sparsity_cond} shows that the dynamic neural network with $4$ horizontal layers simulates the LTI system accurately compared to the numerical solution using the Python routine \texttt{SciPy.signal.lsim}. 

\subsection{State matrix with different kinds of clusters of eigenvalues}
\label{ss:blobs}
The objective of the next numerical example is to demonstrate that a dynamic neural network with all types of horizontal layers can simulate LTI systems accurately and thus validate the implementation.
Each horizontal layer either consists of only first-order neurons, only second-order neurons, or both. 
This depends on the number of real and complex eigenvalues in the corresponding diagonal block of the state matrix. 

\begin{figure}
\begin{center}
\begin{minipage}[b]{0.3\textwidth}
    \centering
    \includegraphics[height=0.7\textwidth, width=0.9\textwidth]{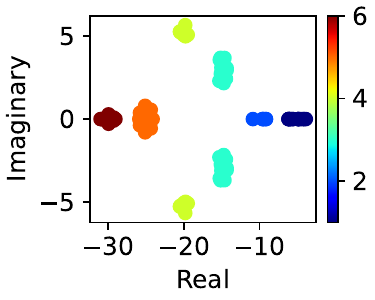}
    \captionof{figure}{Eigenvalue clustering. Colors indicate eigenvalue clusters (Ex 3.3).}
    \label{fig:eval_clustering_blobs}
\end{minipage}\hspace{0.15cm}
\begin{minipage}[b]{0.3\textwidth}
    \centering
    \includegraphics[height=0.7\textwidth, width=0.9\textwidth]{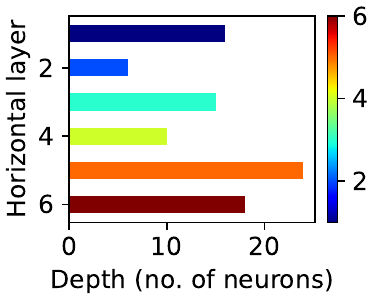}
    \captionof{figure}{DyNN architecture. Colors indicate horizontal layers (Ex 3.3).}
    \label{fig:DyNN_architecture_blobs}
\end{minipage}\hspace{0.15cm}
\begin{minipage}[b]{0.36\textwidth}
    \centering
    \includegraphics[width=0.72\textwidth]{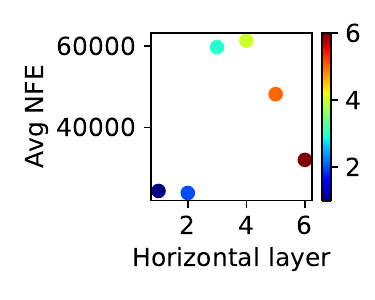} 
    \captionof{figure}{Average NFE count of horizontal layers. Colors indicate horizontal layers (Ex 3.3).}
    \label{fig:n_nfe_blobs}
\end{minipage}
\end{center}
\vspace{-2.3em}
\end{figure}

We construct an LTI system with input, state, and output dimensions $d_i = 10$, $d_h = 134$ and $d_o = 4$. 
The state matrix $\underbar{$A$} \in \mathbb{R}^{134 \times 134}$ is initialized as a block upper triangular matrix with blocks of size $1 \times 1$ or $2 \times 2$, whose eigenvalues are chosen as shown in the \cref{fig:eval_clustering_blobs}. 
To validate the applicability of our algorithm to dense matrices, we define a new coordinate transformation via a random rotation matrix $\mathcal{R}$ and perform a similarity transformation to get a new state-space model that preserves the input-output map of the LTI system and ensures that the new state matrix $\Tilde{A} = \mathcal{R}^{-1} \underbar{A}  \mathcal{R}$ is dense. 
See \cref{app:ps_ex3} for details on the problem setup. 

\begin{figure}
    \centering
    \includegraphics[width=0.7\textwidth]{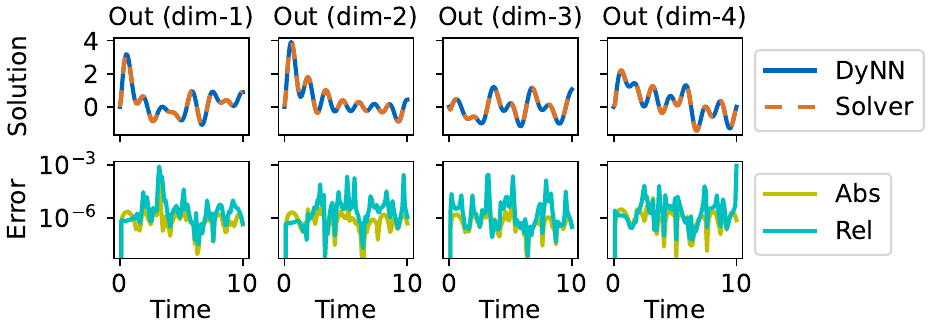}   
    \captionof{figure}{Top panel: Outputs of DyNN and numerical solver.
    Bottom panel: relative and absolute errors between the DyNN output and numerical solution (Ex 3.3).}
    \label{fig:n_sol_error_blobs}
\vspace{-2.2em}
\end{figure}

\cref{fig:eval_clustering_blobs} shows how the eigenvalues are clustered and  
\cref{fig:DyNN_architecture_blobs} shows the corresponding hidden layer architecture of the constructed DyNN. 
Due to the partial decoupling of state dynamics across six diagonal blocks, the NFE count of ODE solvers for neurons in any horizontal layer is independent of the NFE count of neurons in other horizontal layers.
\cref{fig:n_nfe_blobs} illustrates a high variation in the NFE count of neurons averaged over each horizontal layer. 
With six clusters of eigenvalues, the condition number of the transformation matrix  $C_{tr}$ is $11.2$. If seven clusters are chosen, close eigenvalues are forced in different clusters and $C_{tr}$ becomes $7.57 \times 10^{6}$. 
\cref{fig:n_sol_error_blobs} shows that the DyNN with $6$ horizontal layers simulates the LTI system with high accuracy compared to the numerical simulation using a Python routine \texttt{SciPy.signal.lsim}.

\begin{remark}
For simulation results of the convection-diffusion equation, please refer to \ref{app:ss:convection}. 
Secondly, our numerical analysis and experimental results are based on \cref{alg:dynamic neural network}, in which we solve the ODEs of all neurons over the entire time domain. 
In practice, if we solve the ODEs over smaller time intervals instead of the entire time domain (see \cref{alg:dynamic neural network_timesteps}), the inference time can be lower, and the errors compared to the numerical solver are lower (see \cref{sm:timesteps_first}). 
\end{remark}

%% file: sections/conclusions_discussion.tex
\section{Conclusions and Discussion} \label{sec:conclusion}
In this work, we outlined a path toward systematically constructing sparse neural network architectures for modeling dynamical systems.
Starting with the state-space formulation of the LTI system, we derived a mapping from the parameters of the LTI system to the parameters of the continuous-time neural network to compute the latter without gradient-based iterative optimization. 
We introduced a novel paradigm of neural architectures with 'horizontal layers' and demonstrated how enforcing sparsity by using vertical layers may result in highly ill-conditioned transformation matrices and blow up the network weights.
We proved that the numerical error introduced by our continuous-time neural networks is of the same order as the error produced by the ODE solver of each neuron and empirically demonstrated that our networks can accurately simulate general LTI systems.

Gradient-free computation of network parameters implies that black-box and non-differentiable ODE solvers can be used to compute the state of each neuron in the forward pass. 
Since the pre-processing algorithm has the potential to separate slow and fast dynamics across different horizontal layers, the NFE count may exhibit significant variation across horizontal layers. 
This could significantly reduce computational costs during inference by not needing to evaluate the entire state matrix in each function evaluation of the ODE solver. 
Each neuron can even use a different ODE solver.
\subsection{Potential extensions of the current work}
We focus on developing a framework for the principled construction of sparse architectures for general LTI systems, considering the conditioning of transformation matrices.
As a consequence of this, the current pre-processing algorithm involves computationally expensive operations.
For certain special LTI systems, however, one can cheaply construct different sparse architectures by representing the state-space models using special canonical forms.
Moreover, smarter interpolation strategies for dense outputs of neurons, and exploiting parallelization in the forward pass across horizontal layers will lead to faster inference time in applications. 

This work only addresses LTI systems.  
An important future direction is to systematically construct dynamic neural networks for more challenging classes of dynamical systems such as linear parameter-varying systems, quadratic bilinear systems, and, ultimately, more involved non-linear dynamical systems. 

There is a vast literature on Model-Order Reduction (MOR) for dynamical systems \cite{schilders2008model,benner2020model2,benner2020model3}. 
Due to the absence of a systematic technique for constructing neural networks from dynamical systems, there has been a minimal investigation into the theory of MOR for reducing neural networks.
Our work takes an initial stride towards enabling this transfer of knowledge and constructing reduced neural networks.
An interesting extension in light of this is extending adjoint-based methods or designing appropriate differentiable ODE solvers for our architectures. 
One can then compute the architecture and parameters of the reduced network as a starting point and fine-tune the parameters further with gradient-based methods using non-linear activation functions.

%% file: sections/appendix.tex
\section{Constructing Dynamic Neural Networks for general LTI Systems}

\subsection{\textbf{Clustering algorithm}} \label{rem:sm_clustering}
The parameter \texttt{clustering algorithm} in \cref{alg:preprocessing_lti} for grouping eigenvalues of the state matrix $\Tilde{A}$ can be chosen from the myriad clustering algorithms.
These include the well-known k-means algorithm \cite{lloyd1982least}, the spectral clustering algorithm \cite{von2007tutorial}, and others.
These algorithms and many more are implemented in the Python package \texttt{scikit-learn} \cite{scikit-learn}. 
For each cluster of eigenvalues, we identify the eigenvalue with the largest real part and sort clusters in descending order based on these, i.e., the cluster containing the eigenvalue with the maximum real part is numbered $1$. The one having the eigenvalue with the lowest real part is numbered $L$ if there are $L$ clusters. 
This can be skipped in practice, but it ensures that the algorithm is deterministic. 
Within each cluster of eigenvalues, we order the eigenvalues according to the absolute value of the real part in ascending order. 
For a cluster with real and complex eigenvalues, this internal ordering ensures that real eigenvalues are placed first (on the diagonal blocks of the transformed state matrix $A$). 
This results in a suitable sparsity pattern of the diagonal block (as shown in the matrix on the right in equation \eqref{eq:sparsity_patterns}), which is exploited in \cref{th:parameter_maps}.  

\subsection{\textbf{Arithmetic complexity of the pre-processing \cref{alg:preprocessing_lti}}} \label{app:arithmatic_complexity}
For a given state-matrix $A \in \mathbb{R}^{n \times n}$, the real Schur decomposition requires $\mathcal{O}(n^3)$ as $n \to \infty$ floating point operations \cite{golub2013matrix}. 
Ordering $k$ eigenvalues in a Schur form has an arithmetic complexity of  $\mathcal{O}(kn^2)$ \cite{granat2009parallel}. 
The exact cost depends on the distribution of the eigenvalues over the diagonal of the Schur form. 
Block-diagonalization of the state matrix involves solving Sylvester equations. 
If two diagonal blocks of the transformed state matrix have dimensions $\mathbb{R}^{m \times m}$ and $\mathbb{R}^{n \times n}$, respectively, the Bartels-Stewart algorithm requires 
$\mathcal{O}(m^3 + n^3)$ floating point operations to solve the Sylvester equation
\cite{kirrinnis2001fast}. 
However, the number of flops required by the full for block-diagonalization using the Bartels-Stewart Algorithm is a complicated function of the block sizes \cite{golub2013matrix} and, though not common, can have complexity up to $\mathcal{O}(n^4)$ \cite{bavely1979algorithm}. 
Thus, the worst arithmetic complexity of our entire algorithm is also $\mathcal{O}(n^4)$. 
\subsection{Theoretical results on mapping from parameters of the LTI system to parameters of the DyNN} \label{sec:app_theory}

\subsubsection{Similarity transformation via permutation}
\begin{lemma}[Similarity transformation via permutation]\label{lemma:permute}
Any matrix $M\in \mathcal{G}_c$ (see \cref{def:sparse_G}) can be transformed via a similarity transformation $ M\mapsto T M T^{-1}$ with the permutation matrix
$T=\begin{bmatrix}
        I_n \otimes \begin{bmatrix}
            1&0
        \end{bmatrix}\\
        I_n \otimes \begin{bmatrix}
            0&1
        \end{bmatrix}
        \end{bmatrix}$ into a block matrix as \begin{align*}
    \left[\begin{array}{cc|cc|c|cc}
        a_{11}  &b_{11}     &a_{12}     &b_{12}    &\cdots     &a_{1n}  &{b_{1n}}\\
        c_{11}  &d_{11}     &c_{12}     &d_{12}    &\cdots     &c_{1n}  &{d_{1n}}\\
        \hline 
        0       &0          &a_{22}     &b_{22}    &\cdots     &a_{2n}  &{b_{2n}}\\
        0       &0          &c_{22}     &d_{22}    &\cdots     &c_{2n}  &{d_{2n}}\\
        \hline 
        0       &0          &0          &0          &\ddots     &\vdots &\vdots\\
        0       &0          &0          &0          &\ddots     &\vdots &\vdots\\
        \hline
        0       &0          &0          &0         &0           &a_{nn}  &{b_{nn}}\\
        0       &0          &0          &0         &0           &c_{nn}  &{d_{nn}}\\
    \end{array}\right] \mapsto \begin{bmatrix}
        A&B\\C&D
    \end{bmatrix},
\end{align*}
where the blocks $A,B,C$ and $D$ are upper triangular and 
\begin{align*}
    \begin{bmatrix}
        A&B\\C&D
    \end{bmatrix} = \left[\begin{array}{cccc|cccc}
    a_{11}      &a_{12}     &\cdots     &a_{1n}     &b_{11}     &b_{12}     &\cdots &b_{1n}\\
    0           &a_{22}     &\cdots     &a_{2n}     &0          &b_{22}     &\cdots &b_{2n}\\
    0           &0          &\ddots     &\vdots     &0          &0          &\ddots &\vdots\\
    0           &0          &0          &a_{nn}     &0          &0          &0      &b_{nn}\\
    \hline
    c_{11}      &c_{12}     &\cdots     &c_{1n}     &d_{11}     &d_{12}     &\cdots &d_{1n}\\
    0           &c_{22}     &\cdots     &c_{2n}     &0          &d_{22}     &\cdots &d_{2n}\\
    0           &0          &\ddots     &\vdots     &0          &0          &\ddots &\vdots\\
    0           &0          &0          &c_{nn}     &0          &0          &0      &d_{nn}\\
    \end{array}\right]
\end{align*}
for any $\{a_{ij},b_{ij},c_{ij},d_{ij}\}$ for any $i,j \in \{1,\cdots,n\}$.
\end{lemma}

\begin{figure}[htbp]
    \centering
    \includegraphics[width=\textwidth]{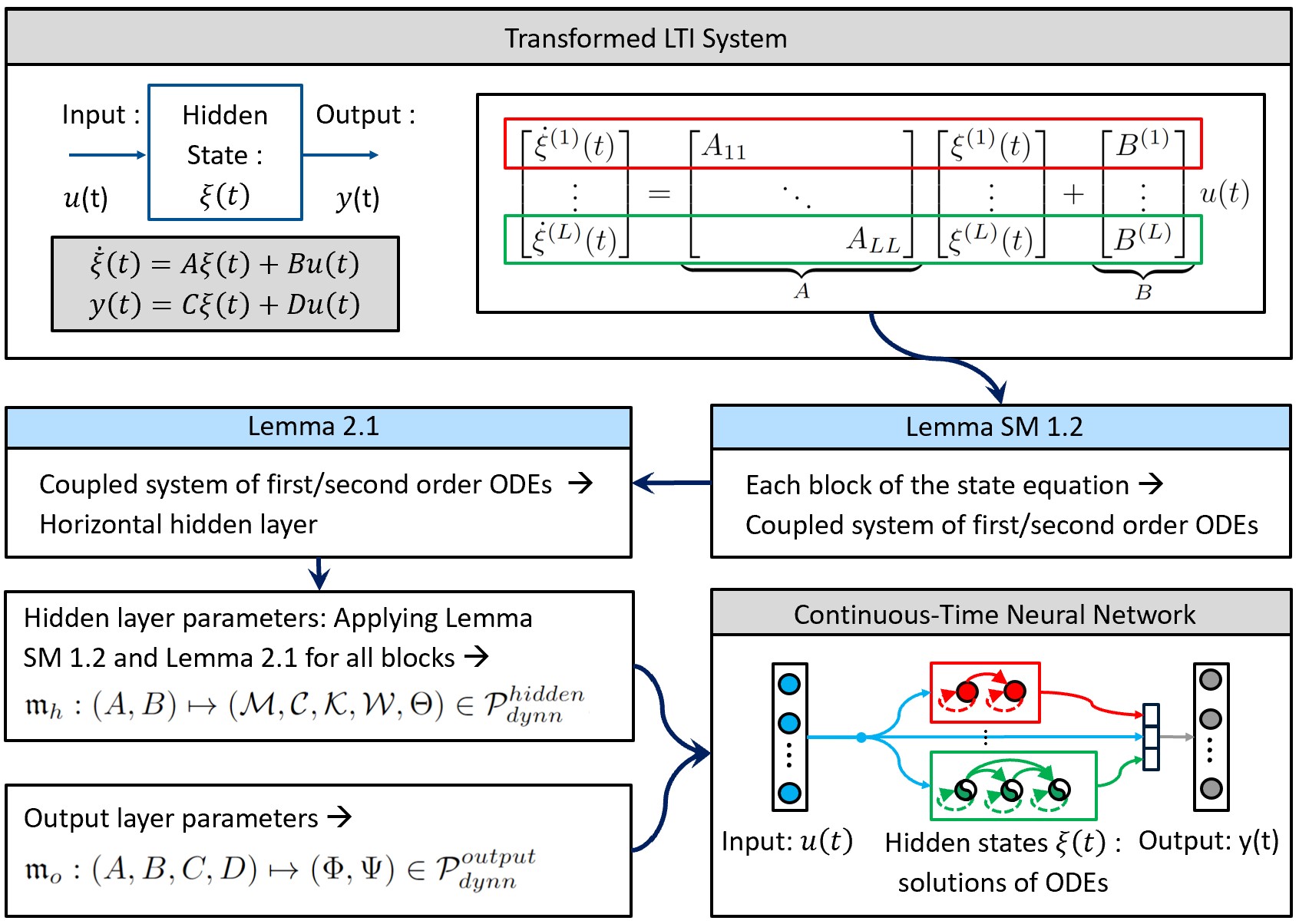}   
        \captionof{figure}{Illustration of how different theoretical results are interconnected and used in \cref{th:parameter_maps} describing the mapping from parameters of transformed LTI system to parameters of the dynamic neural network.}
    \label{fig:theory_workflow}
\vspace{-2em}
\end{figure}

\subsubsection{Proof of \cref{lemma: overall_dynamics_DyNN}}\label{app_lemma: overall_dynamics_DyNN}
\begin{proof}[Proof of \cref{lemma: overall_dynamics_DyNN}]
    We prove the theorem by constructing the map $\mathfrak{n}^{(l)}_{dynn}$.
    For given parameters $\left(\mathcal{M}^{(l)},\mathcal{C}^{(l)},\mathcal{K}^{(l)},\mathcal{W}^{(l)}\right)$ of layer $l$,
    we have from \cref{def_iso_dynn} and \cref{def_io_neuron}, for $i \in \{1, \hdots, n_l\}$, that 
    \begin{align}\label{eq:single_neuro_ode}
        m^{(l)}_i \; \ddot{\xi}^{(l)}_{i}(t) + c^{(l)}_i \; \dot{\xi}^{(l)}_{i} + k^{(l)}_i\xi^{(l)}_{i}(t) &= w_i^{(l)} u_i^{(l)}(t).
    \end{align}
    We partition $w_i^{(l)}$ according to the partition of $u_i^{(l)}$ as defined in equation \eqref{eq:neuron_map} and define $e_i^{(l)},v_i^{(l)}$ and $k_{i,j}^{(l)},c_{i,j}^{(l)}$ for $j \in \{i+1,\cdots,n_l\}$ as the sub-partitions of $w_i^{(l)}$ as 
    \begin{align*}
        w_i^{(l)}=\left[\begin{array}{c|c|c|c|c}
             e_i^{(l)}& v_i^{(l)} & -k_{i,i+1}^{(l)}\,-c_{i,i+1}^{(l)}& \cdots&-k_{i,n_l}^{(l)}\, -c_{i,n_l}^{(l)}
        \end{array}\right].
    \end{align*}
    Substituting this in \eqref{eq:single_neuro_ode} and using \cref{def_iso_dynn} along with the notation $k_{i,i}^{(l)}:=k_i^{(l)}$, $c_{i,i}^{(l)}:=c_i^{(l)}$, we have
    \begin{align}
        &m^{(l)}_i \; \ddot{\xi}^{(l)}_{i}(t) + \sum_{j=i}^{n_l} c_{i,j}^{(l)} \dot{\xi}_j^{(l)}(t)+ \sum_{j=i}^{n_l} k_{i,j}^{(l)} \xi_j^{(l)}(t) = e_i^{(l)}u(t)+v_i^{(l)}(t)\dot{u}(t)
    \end{align}
    for $l\in\{1,\cdots,L\}$.
    Writing these equations in matrix form, we have
    \begin{equation*}
        M^{(l)}\Ddot{\xi}^{(l)}(t)+C^{(l)}\dot{\xi}^{(l)}(t)+K^{(l)}\xi^{(l)}(t)=E^{(l)}u(t)+V^{(1)}\dot{u}(t) \quad \forall t \in \Omega,
    \end{equation*}
    where $ M^{(l)}$, $C^{(l)}$, $ K^{(l)}$ $E^{(l)}$ and $V^{(l)}$ are as defined in equation \eqref{eq:m_h_map} in \cref{ssec:n_dynn}.
    Note that as the matrices $\left(M^{(l)}, C^{(l)}, K^{(l)}, E^{(l)}, V^{(l)}\right)$ defined above belong to $\mathcal{S}_{n_l,d_i}$ (see \cref{def:matrices_2nd_order_formalism}), equations \eqref{eq:m_h_map} together with the definitions of $e_i^{(l)},v_i^{(l)}$ and $k_{i,j}^{(l)},c_{i,j}^{(l)}$ for $j \in \{i+1,\cdots,n_l\}$ as entries of $w_i^{(l)}$ together define the sought map $\mathfrak{n}^{(l)}_{dynn}$.
    Conversely, observe that if the matrices $\left(M^{(l)}, C^{(l)}, K^{(l)}, E^{(l)}, V^{(l)}\right)\in \mathcal{S}_{n_l,d_i}$ are given, we can construct
    $w_i^{l}$ (and therefore the tuple $\mathcal{W}^{(l)}$) from $ E^{(l)}, V^{(l)}$ and the off-diagonal entries of $C^{(l)}, K^{(l)}$.
    Finally, $\mathcal{M}^{(l)},\mathcal{C}^{(l)},\mathcal{K}^{(l)}$ can be constructed from the diagonal entries of $M^{(l)},C^{(l)}, K^{(l)}$.
    This completes the proof.
\end{proof}

\subsubsection{First and/or second order dynamics of an LTI system}
\begin{lemma}[First and/or second order dynamics of an LTI system] \label{lemma:m_lti_to_second_order}
        For some non-negative integers $k_r, k_c$ and $d_i \in \mathbb{N}$, let $\mathcal{A} \in \mathcal{G} \subset \mathbb{R}^{(k_r+2k_c) \times (k_r+2k_c)}$ have $k_r$ real eigenvalues and $k_c$ pairs of complex eigenvalues, 
        and $\mathcal{B} \in \mathbb{R}^{(k_r+2k_c) \times d_i}$ .
	Let the input $u \in \mathcal{C}^1(\Omega)^{d_i}$ and state $x\in \mathcal{C}^2(\Omega)^{(k_r+2k_c)}$ satisfy the linear differential equation
	\begin{align*}
		\dot{x}=\mathcal{A}x+\mathcal{B}u, \quad x(0)=0.
	\end{align*}
    The mappings
	\begin{align*}
		\mathfrak{m}_{lti}&: (\mathcal{A}, \mathcal{B}) \mapsto (M, C, K, E, V) \in \mathcal{S}_{k_r+k_c} \quad \textnormal{(see \cref{def:matrices_2nd_order_formalism})},\\
		\mathfrak{m}_{\eta}&: (\mathcal{A}, \mathcal{B}) \mapsto (W, Q, Z) 
	\end{align*}
	as described in \cref{ssec:m_eta_and_m_lti} can be constructed such that the new variables $\xi(t)\in\mathbb{R}^{k_r+k_c}$, $\eta(t)\in\mathbb{R}^{k_c}$, $\xi_r(t)\in\mathbb{R}^{k_r}$ and $\xi_c(t)\in\mathbb{R}^{k_c}$ defined as 
	\begin{align*}
		\left[\begin{array}{c}
			\xi(t)\\
			\hline
			\eta(t)
		\end{array}\right] \coloneqq \left[\begin{array}{c}
		\xi_r(t)\\
		\xi_c(t)\\
		\hline
		\eta(t)
		\end{array}\right]=
		\left[\begin{array}{cc}
			I_{k_r} & 0\\
			0 &	I_{k_c} \otimes \begin{bmatrix} 1&0 \end{bmatrix}\\
			\hline
			0 &	I_{k_c} \otimes \begin{bmatrix} 0&1 \end{bmatrix}
		\end{array}
		\right]
				x(t)
	\end{align*}
	satisfy 
	\begin{align}
		&M\Ddot{\xi}(t)+C\dot{\xi}(t)+K\xi(t)=Eu(t)+V\dot{u}(t),\\
		&\eta(t)=W \, \xi_c(t) + Q \dot{\xi}_c(t) + Z u(t),
	\end{align}
	for all $t \in \Omega$ with $\xi(0)=0$, $\eta(0)=0$.
    Furthermore, the matrices $W, Q \in \mathbb{R}^{k_c \times k_c}$ are upper-triangular, i.e., $W_{ij} = 0, Q_{ij} = 0$ for $i > j$ and $Z \in  \mathbb{R}^{k_c \times d_i}$.
\end{lemma}
\begin{proof}
	Since $\mathcal{A} \in \mathcal{G}$, it has the form
	\begin{align*}
		\mathcal{A} =\begin{bmatrix}
			\mathcal{A}_{r} & \mathcal{A}_{rc}\\
			0 & \mathcal{A}_{c}
		\end{bmatrix},
	\end{align*}
	where $\mathcal{A}_r \in \mathcal{G}_r \subset \mathbb{R}^{k_r \times k_r}$ and $\mathcal{A}_c \in \mathcal{G}_c \subset \mathbb{R}^{2k_c \times 2k_c}$ (see \cref{def:sparse_G}). 
    Note that the matrix $\mathcal{A}_{rc}$ in the notation described here does not represent (block-) row $r$ and (block-) column $c$ of the matrix $\mathcal{A}$, it represents the first $k_r$ rows and last $2k_c$ columns of $\mathcal{A}$. 
	For convenience, let 
	\begin{align*}
			T&=\begin{bmatrix}
			I_{k_c} \otimes \begin{bmatrix}
				1&0
			\end{bmatrix}\\
			I_{k_c} \otimes \begin{bmatrix}
				0&1
			\end{bmatrix}
		\end{bmatrix},\\
		P&=\begin{bmatrix}
			I_{k_r} & 0\\ 0 & T
		\end{bmatrix}.
	\end{align*}
	Since $P$ is a permutation matrix, i.e., $P^{-1}=P^T$,
	we get that
	\begin{align*}
		\begin{bmatrix}
			\dot{\xi}(t) \\
			\dot{\eta}(t)
		\end{bmatrix}=		
		P\dot{x}(t)=P\mathcal{A}x(t)+P\mathcal{B}u(t) &=P\mathcal{A}P^T\begin{bmatrix}
		\xi(t) \\
		\eta(t)
		\end{bmatrix}+P\mathcal{B}u(t) \nonumber\\
		&=\begin{bmatrix}
			\mathcal{A}_{r} & \mathcal{A}_{rc}T^T\\
			0 & T\mathcal{A}_{c}T^T
		\end{bmatrix}
		\begin{bmatrix}
			\xi(t) \\
			\eta(t)
		\end{bmatrix}
		+
		P\mathcal{B} u(t).
	\end{align*}
	Since $\mathcal{A}_c \in \mathcal{G}_c$, we can apply \cref{lemma:permute} to see that $T\mathcal{A}_{c}T^T$ is a block $2\times 2$ matrix with each block being upper-triangular.
	Thus, we get that  
	\begin{align}\label{eq:xi_eta_sys}
		\begin{bmatrix}
			\dot{\xi}_r(t)\\
			\dot{\xi}_c(t)\\
			\dot{\eta}(t)
		\end{bmatrix} &=\begin{bmatrix}
			\mathcal{A}_{11} & \mathcal{A}_{12} & \mathcal{A}_{13}\\
			0 & \mathcal{A}_{22} & \mathcal{A}_{23} \\
			0 & \mathcal{A}_{32} & \mathcal{A}_{33} 
		\end{bmatrix}
		\begin{bmatrix}
			{\xi}_r(t)\\
			{\xi}_c(t)\\
			{\eta}(t)
		\end{bmatrix}+\begin{bmatrix}
		\mathcal{B}_1\\
		\mathcal{B}_2\\
		\mathcal{B}_3
		\end{bmatrix}u(t)
	\end{align} 
	where $\mathcal{A}_{11}=\mathcal{A}_r$ is upper triangular and the blocks $\mathcal{A}_{22},\mathcal{A}_{23},\mathcal{A}_{32},\mathcal{A}_{33}$ are all upper-triangular matrices because of \cref{lemma:permute}.
Furthermore, since the super-diagonal of $\mathcal{A}_c$ is transformed to the main diagonal of $\mathcal{A}_{23}$ (see \cref{lemma:permute}), $\mathcal{A}_{23}$ is an invertible matrix owing to \cref{def:sparse_G} (condition of non-zero super-diagonal entries of all $2 \times 2$ diagonal blocks). 
Solving the second equation of \eqref{eq:xi_eta_sys} for $\eta(t)$, we get
	\begin{align} \label{eq:x_o}
		\eta(t) &=   \mathcal{A}_{23}^{-1} \left(\dot{\xi}_c(t) - \mathcal{A}_{22} \xi_c(t) -\mathcal{B}_2 u(t) \right) \nonumber \\
		&= \underbrace{\left(-\mathcal{A}_{23}^{-1}\mathcal{A}_{22}\right)}_{W} \xi_c(t) + \underbrace{\mathcal{A}_{23}^{-1}}_{Q}\dot{\xi}_c(t) + \underbrace{\left(-\mathcal{A}_{23}^{-1}\mathcal{B}_2\right)}_{Z} u(t).
	\end{align}
	The sought map $\mathfrak{m}_{\eta}$ is thus found with the above definitions of $(W,Q,Z)$. 
Differentiating the second equation of \eqref{eq:xi_eta_sys} with respect to time, we get 
	\begin{align*}
		\ddot{\xi}_c(t) &= \mathcal{A}_{22} \, \dot{\xi}_c(t) + \mathcal{A}_{23} \dot{\eta}(t) + \mathcal{B}_2 \dot{u}(t) \\
		&= \mathcal{A}_{22}\dot{\xi}_c(t) + \mathcal{A}_{23} \underbrace{\left( \mathcal{A}_{32}\xi_c(t) + \mathcal{A}_{33}\eta(t) + \mathcal{B}_3 u(t) \right)}_{\dot{\eta}(t)} + \mathcal{B}_2 \dot{u}(t)\\
		&= \mathcal{A}_{22}\dot{\xi}_c(t) + \mathcal{A}_{23} \mathcal{A}_{32}\xi_c(t) + \mathcal{A}_{23}\mathcal{A}_{33}\eta(t) + \mathcal{A}_{23}\mathcal{B}_3 u(t)+ \mathcal{B}_2 \dot{u}(t)
	\end{align*}
	where we have substituted $\dot{\eta}(t)$ from the third equation of \eqref{eq:xi_eta_sys}.
	Substituting $\eta(t)$ from equation \eqref{eq:x_o}, we get
	\begin{align} \label{eq:complex_sos}
		\ddot{\xi}_c(t) &= \mathcal{A}_{22}\dot{\xi}_c(t) + \mathcal{A}_{23} \mathcal{A}_{32}\xi_c(t) + \mathcal{A}_{23}\mathcal{A}_{33}\overbrace{\left(W \xi_c(t) +Q\dot{\xi}_c(t) +Z u(t)\right)}^{\eta(t)} \nonumber \\ &\quad+ \mathcal{A}_{23}\mathcal{B}_3 u(t) + \mathcal{B}_2 \dot{u}(t) \nonumber \\
	&= \underbrace{\left(\mathcal{A}_{22}+\mathcal{A}_{23}\mathcal{A}_{33}Q \right)}_{-C_c}\dot{\xi}_c(t) + \underbrace{\left(\mathcal{A}_{23} \mathcal{A}_{32} +\mathcal{A}_{23}\mathcal{A}_{33}W\right)}_{-K_c}\xi_c(t) \nonumber \\ 
            &\quad+ \underbrace{\left( \mathcal{A}_{23}\mathcal{A}_{33}Z 
            + \mathcal{A}_{23}\mathcal{B}_3\right)}_{E_c}u(t) + \underbrace{\mathcal{B}_2}_{V_c} \dot{u}(t).
	\end{align}
	Finally, substituting $\eta(t)$ from equation \eqref{eq:x_o} into the first equation of \eqref{eq:xi_eta_sys}, we get
	\begin{align}\label{eq:real_sos}
		\dot{\xi}_r(t)&=\mathcal{A}_{11} \xi_r(t) + \mathcal{A}_{12} \xi_c(t) + \mathcal{A}_{13}\eta(t) + \mathcal{B}_1u(t)\nonumber \\
					  &=\mathcal{A}_{11} \xi_r(t) + \mathcal{A}_{12} \xi_c(t) + \mathcal{A}_{13}\underbrace{\left(W \xi_c(t) +Q\dot{\xi}_c(t) +Z u(t)\right)}_{\eta(t)}+\mathcal{B}_1u(t) \nonumber \\
					  &=\underbrace{\mathcal{A}_{11}}_{-K_r}\xi_r(t) + \underbrace{\left(\mathcal{A}_{12}+\mathcal{A}_{13} W\right)}_{-K_{rc}}\xi_c(t) + \underbrace{\mathcal{A}_{13}Q}_{-C_{rc}}\dot{\xi}_c(t)+\underbrace{\left(\mathcal{A}_{13}Z+\mathcal{B}_1\right)}_{E_r}u(t).
	\end{align}	
	Putting together \eqref{eq:complex_sos} and \eqref{eq:real_sos}, we get that
	\begin{align*}
		&\underbrace{\begin{bmatrix}
				0&0\\0& I_{k_c}
		\end{bmatrix}}_M\begin{bmatrix}
			\Ddot{\xi}_r(t)\\
			\Ddot{\xi}_c(t)
		\end{bmatrix}+
		\underbrace{\begin{bmatrix}
				I_{k_r} &C_{rc}\\0& C_c
		\end{bmatrix}}_C\begin{bmatrix}
		\dot{\xi}_r(t)\\
		\dot{\xi}_c(t)
		\end{bmatrix}+
		\underbrace{\begin{bmatrix}
				K_r&K_{rc}\\0& K_c
		\end{bmatrix}}_K\begin{bmatrix}
		{\xi}_r(t)\\
		{\xi}_c(t)
		\end{bmatrix}=
		\underbrace{\begin{bmatrix}
				E_r\\ E_c
		\end{bmatrix}}_E
		u(t)+
		\underbrace{\begin{bmatrix}
				0\\ V_c
		\end{bmatrix}}_V\dot{u}(t).
	\end{align*}
	Since the inverse of an upper-triangular matrix is upper-triangular and the product of upper-triangular matrices is upper-triangular, the matrices $C_c,K_r,K_c$ defined above are all upper-triangular matrices and therefore $(M, C, K, E, V) \in \mathcal{S}_{k_r+k_c}$ thereby defining $\mathfrak{m}_{lti}$ and completing the proof.
\end{proof}

\subsection{Collection of all mappings}
\label{sec:all_mappings}
\subsubsection{\texorpdfstring{Mappings $\mathfrak{m}_{\eta}$ and $\mathfrak{m}_{lti}$:}{Mappings $\mathfrak{m}_n$ and $\mathfrak{m}_{lti}$}}\label{ssec:m_eta_and_m_lti}
We assume $k_{r}$, $k_{c}$, $d_{i}$, and matrices $\mathcal{A} \in \mathcal{G}$  $\subset \mathbb{R}^{(k_{r} + 2k_{c}) \times (k_{r} + 2k_{c})}$ and $\mathcal{B} \in \mathbb{R}^{(k_{r} + 2k_{c}) \times d_{i}}$ are given.
We will next describe the mappings 
\begin{align*}
    \mathfrak{m}_{\eta}&: (\mathcal{A}, \mathcal{B}) \mapsto (W, Q, Z),\\
    \mathfrak{m}_{lti}&: (\mathcal{A}, \mathcal{B}) \mapsto (M, C, K, E, V) \in \mathcal{S}_{k_r+k_c} \quad \textnormal{(see \cref{def:matrices_2nd_order_formalism})}.
\end{align*}
\begin{subequations}\label{eq:m_eta_app}
We start by partitioning the matrix $ \mathcal{A}$ as 
\begin{align}
   \mathcal{A}=\begin{bmatrix}
			\mathcal{A}_{r} & \mathcal{A}_{rc}\\
			0 & \mathcal{A}_{c}
		\end{bmatrix} 
\end{align}
with $\mathcal{A}_r \in \mathcal{G}_r \subset \mathbb{R}^{k_r \times k_r}$, $\mathcal{A}_c \in \mathcal{G}_c \subset \mathbb{R}^{2k_c \times 2k_c}$ (see \cref{def:sparse_G}) and 
define blocks $\mathcal{A}_{ij}$ and $\mathcal{B}_i$ for $i,j \in \{1, 2, 3\}$ and as
\begin{align}    
          \left[
          \begin{array}{c|cc}
             \mathcal{A}_{11}  & \mathcal{A}_{12}  & \mathcal{A}_{13}\\
             \hline
            0              & \mathcal{A}_{22}  & \mathcal{A}_{23}\\  
            0              & \mathcal{A}_{32}  & \mathcal{A}_{33}\\ 
          \end{array}
          \right]
          &=
          \left[
          \begin{array}{c|c}
               \mathcal{A}_{r} & \mathcal{A}_{rc}T^T\\
               \hline
                0 & T\mathcal{A}_{c}T^T
          \end{array}
          \right]
          , \quad 
          \left[
          \begin{array}{c}
               \mathcal{B}_1\\
               \hline
        		\mathcal{B}_2\\
        		\mathcal{B}_3
          \end{array}
          \right]=P\mathcal{B},
    \end{align}
    where the blocks $ \mathcal{A}_{22}, \mathcal{A}_{23}, \mathcal{A}_{32},\mathcal{A}_{33} \in \mathbb{R}^{k_c \times k_c}$
    and
    \begin{align}
            T&=\begin{bmatrix}
			I_{k_c} \otimes \begin{bmatrix}
				1&0
			\end{bmatrix}\\
			I_{k_c} \otimes \begin{bmatrix}
				0&1
			\end{bmatrix}
		\end{bmatrix}, \quad
		P=\begin{bmatrix}
			I_{k_r} & 0\\ 0 & T
		\end{bmatrix}.
  \end{align}
  Finally, the image $(W, Q, Z)$ of $(\mathcal{\mathcal{A}}, \mathcal{B})$ under the map $\mathfrak{m}_{\eta}$ is given by
  \begin{equation}
        W=-\mathcal{A}_{23}^{-1}\mathcal{A}_{22}, \quad Q=\mathcal{A}_{23}^{-1}, \quad Z=-\mathcal{A}_{23}^{-1}\mathcal{B}_2.
    \end{equation}
    \end{subequations}
    We next define the following matrices. 
\begin{subequations} \label{eq:m_lti_app}
    \begin{align}		
        C_{rc}&=-\mathcal{A}_{13}Q,\,  C_c=-\left( \mathcal{A}_{22}+\mathcal{A}_{23}\mathcal{A}_{33}Q \right),  \\
        K_r&=-\mathcal{A}_{11}, \, K_{rc}=-\left(\mathcal{A}_{12}+\mathcal{A}_{13} W\right),\,
        K_c=-\left(\mathcal{A}_{23} \mathcal{A}_{32} +\mathcal{A}_{23}\mathcal{A}_{33}W\right), \\
        E_r&=\left( \mathcal{A}_{13}Z+\mathcal{B}_1\right), \, E_c=\left( \mathcal{A}_{23}\mathcal{A}_{33}Z+ \mathcal{A}_{23}\mathcal{B}_3\right), \\
        V_c&=\mathcal{B}_2.
    \end{align}
Finally, the image $(M, C, K, E, V) \in \mathcal{S}_{k_r+k_c}$ of $(\mathcal{A}, \mathcal{B})$ under the map $\mathfrak{m}_{lti}$ is given by
    \begin{align}
        M&=\begin{bmatrix}
				0&0\\0& I_{k_c}
		\end{bmatrix},
  C=\begin{bmatrix}
				I_{k_r}&C_{rc}\\0& C_c
		\end{bmatrix},
  K=\begin{bmatrix}
				K_r&K_{rc}\\0& K_c
		\end{bmatrix},\\
  E&=\begin{bmatrix}
				E_r\\ E_c
		\end{bmatrix},
  V=\begin{bmatrix}
				0\\ V_c
		\end{bmatrix}.
    \end{align}
\end{subequations}

\subsubsection{Mappings $\mathfrak{n}_{dynn}^{(l)}$ and $[\mathfrak{n}_{dynn}^{(l)}]^{-1}$}\label{ssec:n_dynn}
We next describe the bijective mapping
\begin{subequations}\label{eq:m_h_map_app}
\begin{align} \label{eq:nl_dynn}
\mathfrak{n}_{dynn}^{(l)}:\left(\mathcal{M}^{(l)},\mathcal{C}^{(l)},\mathcal{K}^{(l)},\mathcal{W}^{(l)}\right) \mapsto \left(M^{(l)}, C^{(l)}, K^{(l)}, E^{(l)}, V^{(l)}\right).   
\end{align}
First note that row $i$ of $\mathcal{W}^{(l)}$ is composed of $w_i^{(l)}$ (and similarly $\mathcal{M}^{(l)},\mathcal{C}^{(l)},\mathcal{K}^{(l)}$).
Next, we partition $w_i^{(l)}$ as 
    \begin{equation}\label{eq:w_partition}
        w_i^{(l)} = \left[\begin{array}{c|c|c|c|c}
             e_i^{(l)}& v_i^{(l)} & -k_{i,i+1}^{(l)}\,-c_{i,i+1}^{(l)}& \cdots&-k_{i,n_l}^{(l)}\, -c_{i,n_l}^{(l)}
        \end{array}\right]
    \end{equation}
    to define $e_i^{(l)}\in \mathbb{R}^{1\times d_i}$, $v_i^{(l)}\in \mathbb{R}^{1\times d_i}$, $k_{i,j}^{(l)},c_{i,j}^{(l)}\in\mathbb{R}$ for $i\in\{1,\cdots,n_l\}$ and $j\in\{i+1,n_l\}$.
    Additionally, let $k_{i,i}^{(l)}:=k_i^{(l)}$, $c_{i,i}^{(l)}:=c_i^{(l)}$ for all $i\in\{1,\cdots,n_l\}$. 
    Finally, define the image $\left(M^{(l)}, C^{(l)}, K^{(l)}, E^{(l)}, V^{(l)}\right)$ of $\left(\mathcal{M}^{(l)},\mathcal{C}^{(l)},\mathcal{K}^{(l)},\mathcal{W}^{(l)}\right)$ under the map $\mathfrak{n}_{dynn}^{(l)}$ as 
    \begin{align}\label{eq:m_h_map}
        \begin{split}
            M^{(l)}&=\begin{bmatrix}
            m_1^{(l)} & & &\\
            & m_1^{(l)} & &\\
            & & \ddots &\\
            & & & m_{n_l}^{(l)}\\
        \end{bmatrix}, \quad 
        C^{(l)}=\begin{bmatrix}
            c_{1,1}^{(l)} &c_{1,2}^{(l)} &\cdots &c_{1,n_l}^{(l)}\\
            & c_{2,2}^{(l)} & &\\
            & & \ddots &\vdots\\
            & & & c_{n_l,n_l}^{(l)}\\
        \end{bmatrix}, \\
        K^{(l)}&=\begin{bmatrix}
            k_{1,1}^{(l)} &k_{1,2}^{(l)} &\cdots &k_{1,n_l}^{(l)}\\
            & k_{2,2}^{(l)} & &\\
            & & \ddots &\vdots\\
            & & & k_{n_l,n_l}^{(l)}\\
        \end{bmatrix}, \quad
        E^{(l)}=\begin{bmatrix}
          e_1^{(l)}\\e_2^{(l)}\\ \vdots \\ e_{n_l}^{(l)}
        \end{bmatrix}, \quad
        V^{(l)}=\begin{bmatrix}
          v_1^{(l)}\\v_2^{(l)}\\ \vdots \\ v_{n_l}^{(l)}
        \end{bmatrix}.
        \end{split}
    \end{align}
\end{subequations}

For the inverse map $[\mathfrak{n}_{dynn}^{(l)}]^{-1}$, note that we can read off the elements $e_i^{(l)}\in \mathbb{R}^{1\times d_i}$, $v_i^{(l)}\in \mathbb{R}^{1\times d_i}$, $m_i^{(l)},k_{i,j}^{(l)},c_{i,j}^{(l)}\in\mathbb{R}$ for $i\in\{1,\cdots,n_l\}$ and $j\in\{i,\cdots,n_l\}$ from given matrices $\left(M^{(l)}, C^{(l)}, K^{(l)}, E^{(l)}, V^{(l)}\right)$ as in equation \eqref{eq:m_h_map}.
The image $\left(\mathcal{M}^{(l)},\mathcal{C}^{(l)},\mathcal{K}^{(l)},\mathcal{W}^{(l)}\right)$ of $\left(M^{(l)}, C^{(l)}, K^{(l)}, E^{(l)}, V^{(l)}\right)$ under the inverse map $[\mathfrak{n}_{dynn}^{(l)}]^{-1}$ is then given by setting $w_i^{(l)}$ as in equation \eqref{eq:w_partition}, $k_{i,i}^{(l)}:=k_i^{(l)}$ and $c_{i,i}^{(l)}:=c_i^{(l)}$.

\subsubsection{Mapping $\mathfrak{m}_h$} \label{ssec:m_h}
We next describe the mapping 
\begin{subequations}
\begin{align} \label{eq:m_h_app}
\mathfrak{m}_h:\mathcal{P}_{lti}^{state} \ni \left(A,B\right) \mapsto \left(\mathcal{M},\mathcal{C},\mathcal{K},\mathcal{W}\right) \in \mathcal{P}_{dynn}^{hidden}.  
\end{align}
Since $\left(A, B\right)\in \mathcal{P}_{lti}^{state}$, $A$ is a block-diagonal matrix which is  partitioned together with the appropriate partitioning of $B$ as
\begin{align} \label{eq:AB_partitioning}
    A=\begin{bmatrix}
        A^{(1)}& & & \\
         &A^{(2)}& & \\
         &&\ddots&\\
         &&&A^{(L)}
    \end{bmatrix},\quad
    B=\begin{bmatrix}
        B^{(1)}\\
        B^{(2)}\\
        \vdots\\
         B^{(L)}
    \end{bmatrix},
\end{align}
where $A^{(l)}\in\mathbb{R}^{d_l\times d_l}, B^{(l)}\in \mathbb{R}^{d_l \times d_i}$.
For $l \in \{1, 2, \hdots, L \}$, we construct tuples $(\mathcal{M}^{(l)}$, $\mathcal{C}^{(l)}$, $\mathcal{K}^{(l)}$, $\mathcal{W}^{(l)})$ as
\begin{align}
    [\mathfrak{n}_{dynn}^{(l)}]^{-1} \circ \mathfrak{m}_{lti}:\left(A^{(l)},B^{(l)}\right) \mapsto \left(\mathcal{M}^{(l)},\mathcal{C}^{(l)},\mathcal{K}^{(l)},\mathcal{W}^{(l)}\right).
\end{align}   
\end{subequations}
The tuples $\left(\mathcal{M}^{(l)},\mathcal{C}^{(l)},\mathcal{K}^{(l)},\mathcal{W}^{(l)}\right)$ define the image $\left(\mathcal{M},\mathcal{C},\mathcal{K},\mathcal{W}\right)$ of $\left(A,B\right)$ (see \eqref{eq:parameter_sets_dynn}) under the mapping $\mathfrak{m}_h$.



\subsubsection{Mapping $\mathfrak{m}_o$} \label{ssec:m_0}
\begin{subequations} \label{eq:m_o_app}
We next describe the bijective mapping 
\begin{equation} \label{eq:m_0_app}
\mathfrak{m}_{o}:\left(A,B,C,D\right) \mapsto \left(\Phi,\Psi\right) \in \mathcal{P}_{dynn}^{output}, 
\end{equation}
where $\left(A,B\right) \in \mathcal{P}_{lti}^{state}$, $\left(C,D\right)\in \mathcal{P}_{lti}^{output}$.
We partition the block-diagonal matrix $A$ together with the appropriate partitioning of $B$ as done in equation \eqref{eq:AB_partitioning} so that $A^{(l)}\in\mathbb{R}^{d_l\times d_l}, B^{(l)}\in \mathbb{R}^{d_l \times d_i}$. For $l \in \{1, 2, \hdots, L \}$, we construct tuples $\left(W^{(l)}, Q^{(l)}, Z^{(l)} \right)$ via
\begin{equation}
   \mathfrak{m}_{\eta}: (A^{(l)}, B^{(l)})  \mapsto ( W^{(l)}, Q^{(l)}, Z^{(l)}).
\end{equation}
For $l \in \{1, 2, \hdots, L \}$, and any positive integer $a$, we define the matrices:
    \begin{equation}
        P_{\xi}^{(l)} = \left[\begin{array}{cc}
			I_{k_r^{(l)}} & 0\\
			0 &	I_{k_c^{(l)}} \otimes \begin{bmatrix} 1\\0 \end{bmatrix}
		\end{array}\right], P_{\eta}^{(l)}=\left[\begin{array}{cc}
			0\\
			I_{k_c^{(l)}} \otimes \begin{bmatrix} 1\\0 \end{bmatrix}
		\end{array}\right],
    \end{equation}
    \begin{equation}
         T^{(l)}_a = 
          \begin{bmatrix}
            I_{a} \otimes \begin{bmatrix}
            1&0
            \end{bmatrix}\\
            I_{a} \otimes \begin{bmatrix}
            0&1
            \end{bmatrix}
            \end{bmatrix}
    \end{equation}
We then construct the matrices $\mathcal{F}, \mathcal{Z}$ as
    \begin{align}
        \mathcal{F}^{(l)} &= \begin{bmatrix}
            \left( P_{\xi}^{(l)} +  P_{\eta}^{(l)}\begin{bmatrix}
            0&W^{(l)}
        \end{bmatrix}\right) & \left(P_{\eta}^{(l)}\begin{bmatrix}
            0&Q^{(l)}
        \end{bmatrix}\right)
        \end{bmatrix}T^{(l)}_{n_l},\\
        \mathcal{F} &= \begin{bmatrix}
        \mathcal{F}^{(1)}&&\\
        &\ddots&\\
        &&\mathcal{F}^{(L)}
    \end{bmatrix},\quad
    \mathcal{Z} = \begin{bmatrix}
        P_{\eta}^{(1)}Z^{(1)}(t)\\ \vdots \\ P_{\eta}^{(L)}Z^{(L)}(t)
    \end{bmatrix}.
\end{align}
Finally, for $l \in \{1, 2, \hdots, L\}$ and $i \in \{1,2, \hdots, n_l\}$, construct $\phi_{i}^{(l)} \in \mathbb{R}^{d_o \times 2}$ such that
\begin{align}
    &\left[\begin{array}{c c c|c c c|c|c c c}
            \phi_1^{(1)}&\cdots&\phi_{n_1}^{(1)}&\phi_1^{(2)}&\cdots&\phi_{n_2}^{(2)}&\cdots&\phi_1^{(L)}&\cdots&\phi_{n_L}^{(L)}
        \end{array}\right] = C\mathcal{F},\\
    &\Psi=C\mathcal{Z} + D,
    \end{align}
\end{subequations}
which completes the description of the image $\left(\Phi,\Psi\right)$ of $\left(A,B,C,D\right)$ under the mapping $\mathfrak{m}_{o}$.

\subsection{\textbf{Computation of time-derivatives of the input}} \label{app_rem:comp_time_derivatives}
 If the state matrix has complex eigenvalues, one also needs to compute the time derivative of the input.
 This is a consequence of mapping the two corresponding first-order ODEs to one second-order ODE (see  \cref{lemma:m_lti_to_second_order}), which allows us to treat each pair of eigenvalues as one entity and model the corresponding state dynamics using one second-order neuron. 
 If the input function is differentiable and is available, one can set the time derivative of input as a function handle. 
 If the input function is available only at discrete time points, the time derivative of the input function is computed numerically using a finite difference approximation,  
    $\dot{u}(t) \approx \frac{1}{\Delta t}(u(t+1) - u(t))$, where $u(t+1)$ and $u(t)$ are the inputs at new and old time-steps, respectively, and $\Delta t$ is the time-step size for the current interval.  
If the input function is interpolated as a piece-wise constant function, the state response corresponding to an impulse proportional to the jump in the input is added to the state's output for the subsequent time points. As the input-state map of each neuron is linear, the state response can be calculated separately for the impulse at the initial time point and for the piece-wise constant function over the interval and can be added together.
\subsection{\textbf{Efficient implementation of first-order neurons}} \label{app:eff_imple_firstorder}
In the theory section, we treat a first-order neuron as a special case of a second-order neuron as defined in \cref{def_io_neuron} for convenience, and thus the output of a first-order neuron is given by $y_i^{(l)}(t) =[
    \xi_i^{(l)}(t),
    \dot{\xi}_i^{(l)}(t)]^T$. However, for each first-order neuron, the term $\dot{\xi}_i^{(l)}(t)$ is always multiplied by weights that are set to zero as shown in the proof of \cref{th:parameter_maps} and is thus unnecessary to store. 
For efficient implementation, for each first-order neuron, we define $y_i^{(l)}(t) = \xi_i^{(l)}(t) \in  \mathbb{R}$. The term $\dot{\xi}_i^{(l)}(t)$ and the corresponding weights, which are zero, are not stored.

\section{Numerical examples}
\subsection{Detailed problem setups}
\label{app:num_examples_problem_setups}
\subsubsection{Example 3.1: Diffusion equation (See \cref{ss:diffusion} in the main text)} \label{app:ps_diff}
The domain length $l = 10$ is discretized with $20$ grid points in each dimension. 
The diffusivity $\mathcal{D} = 0.8$. 
Let $t_d$ be a uniform grid in $[0, 10]$ in steps of $0.1$.
We choose a uniform grid in space with a mesh size of $h$, and the Laplacian operator is discretized with second-order finite differences at each grid point $(i, j)$ as $(\frac{\partial^2 T}{\partial x^2} + \frac{\partial^2 T}{\partial y^2})\big|_{i,j} \approx \frac{T_{i+1, j} + T_{i-1, j} + T_{i, j+1} + T_{i, j-1} - 4T_{i, j}}{h^2}$. 
Note that the state matrix $\Tilde{A} \in \mathbb{R}^{400 \times 400}$ is a sparse symmetric matrix.
The other state-space matrices are $\Tilde{B} = \mathcal{I}_{400}$, $\Tilde{C} = \mathcal{I}_{400}$ and $\Tilde{D} = 0$.
We use \texttt{rtol} = $1e^{-10}$, \texttt{atol} = $1e^{-10}$. 
Since the state matrix is unitarily diagonalizable, we expect that the pre-processing \cref{alg:preprocessing_lti} will lead to a transformed state matrix $A$, which is diagonal.   
We use the K-Means clustering algorithm with 400 clusters (equal to the size of the state matrix). 
The code for this example is in the notebook - \texttt{diffusion\_equation\_2d.ipynb}, which contains an exhaustive list of parameter settings and can be used to reproduce figures in the main text. 

\subsubsection{Example 3.2: The reason for horizontal layers (See \cref{ss:horizontal_layers} in the main text)}\label{app:ps_sp_sc}
Note that matrices $\Tilde{B}$, $\Tilde{C}$, and $\Tilde{D}$ could be chosen arbitrarily since $A$ decides the architecture of the constructed DyNN. Here, we sample entries of the input matrix $\Tilde{B} \in \mathbb{R}^{10 \times 10}$ and the output matrix $\Tilde{C} \in \mathbb{R}^{10 \times 10}$ uniformly in $[0, 0.5]$ and sample entries of the feedforward matrix $\Tilde{D} \in \mathbb{R}^{10 \times 10}$ uniformly in $[-0.5,0]$.  
We use \texttt{rtol} = $1e^{-10}$, \texttt{atol} = $1e^{-10}$. We use the K-Means clustering algorithm and vary the number of clusters from one to ten to construct ten different dynamic neural networks. 
Let $t_d$ be a uniform grid in $[0,10]$ in steps of $0.1$. 
The input of the state-space model $u_i(t)$ for $i \in \{1, 2, \hdots, 10\}$ is $ u(i, t) = \sin{(it_d/2)}$ if $t = t_d$ and is interpolated piecewise linearly in time for the points in between.
The code for this example is in the notebook - \texttt{sparsity\_cond\_tradeoff.ipynb}, which contains an exhaustive list of parameter settings and can be used to reproduce the figures in the main text. 

\subsubsection{Example 3.3: State matrix with different kinds of clusters of eigenvalues (See \cref{ss:blobs} in the main text)} \label{app:ps_ex3}
The state matrix $\underbar{$A$} \in \mathbb{R}^{134 \times 134}$ is initialized as a block upper triangular matrix, with all entries above the diagonal blocks sampled from a uniform distribution over $[-0.5,0]$ (an arbitrary choice). 
There are overall $44$ real eigenvalues and $45$ pairs of complex eigenvalues. 
We can control the eigenvalues easily using block-upper triangular matrices with $1 \times 1$ or $2 \times 2$ blocks. 
However, to ensure that our algorithm works for dense state matrices, we define a new coordinate transformation via a random rotation matrix $\mathcal{R}$, drawn from the Haar distribution \cite{stewart1980efficient} so that the new state-space matrices 
\begin{equation} \label{app_eq:transform_abcd}
    \Tilde{A} = \mathcal{R}^{-1} \underbar{A}  \mathcal{R},\quad \Tilde{B} = \mathcal{R}^{-1} \underbar{B},\quad \Tilde{C} = \mathcal{R} \underbar{C}, \quad \Tilde{D} = \underbar{D}
\end{equation}
 are dense but preserve the input-output map of the LTI system and the eigenvalues of the state matrix. 

All complex eigenvalues $\lambda_j = a_j \pm i b_j$ for $j \in {1, 2, \hdots, 45}$ lie on the state matrix as first $45$ diagonal blocks of size $2 \times 2$ in the form $\begin{bmatrix}
    a_j & -b_j \\
    b_j & a_j
\end{bmatrix}$. 
The remaining $44$ real eigenvalues of the state matrix lie on the last $44$ diagonal blocks of size $1 \times 1$.   
We choose the elements of the input matrix $\underbar{$B$} \in \mathbb{R}^{19 \times 10}$ and the output matrix $\underbar{$C$} \in \mathbb{R}^{4 \times 19}$ randomly from a uniform distribution in $[0, 1)$. The elements of feedforward matrix $\underbar{D} \in \mathbb{R}^{4 \times 10}$ are chosen randomly from a uniform distribution in $(-1, 0]$. 
The state space matrices are then transformed with \eqref{app_eq:transform_abcd} to get $(\Tilde{A}, \Tilde{B}, \Tilde{C}, \Tilde{D})$. 
We use \texttt{rtol} = $1e^{-10}$, \texttt{atol} = $1e^{-10}$. We use the K-Means clustering algorithm with 6 clusters.
Let $t_d$ be a uniform grid in $[0,10]$ in steps of $0.1$.
The input of the state-space model $u_i(t)$ for $i \in \{1, 2, \hdots, 10\}$ is $ u(i, t) = \sin{(i t_d / 2)}$ if $t = t_d$ and is interpolated piecewise linearly in time for the points in between. 
The code for this example is in the notebook - \texttt{horizontal\_layers\_all\_types.ipynb}, which contains an exhaustive list of parameter settings and can be used to reproduce the figures in the main text.

\subsection{Additional numerical example: Convection-Diffusion equation}
\label{app:ss:convection}
A two-dimensional transient convection-diffusion equation is
\begin{align} \label{eq:conv_diff}
    \frac{\partial T}{\partial t} = \mathcal{D} \Bigl(\frac{\partial^2 T}{\partial x^2} + \frac{\partial^2 T}{\partial y^2}\Bigr) - v_x \frac{\partial T}{\partial x} - v_y \frac{\partial T}{\partial y} + \mathcal{S}, 
\end{align}
where T is the variable of interest (concentration of species or temperature), $\mathcal{D}$ is diffusivity, $v_x$ and $v_y$ are drift velocities in x and y directions, and $S$ is the source term. 
We interpret this as a system with $S$ as the input and the solution $T$ as the output.
The spatial domain is $[0, 10] \times [0, 9.5]$ with 20 grid points in each dimension. 
The right and left boundaries are periodic. The boundary conditions at the top and bottom boundaries are Dirichlet with $T(x, 0) = T(x, 9.5) = 0$. 

The initial condition $T(x,y,0) = 0$. 
The velocities in x and y dimensions are given by $v_x= 0.6$ and $v_y = 0$, and the diffusivity $\mathcal{D} = 1.4$. 
Let $t_d$ be a uniform grid in $[0,10]$ in steps of $0.1$.
Heat is injected into the system via the source term $\mathcal{S}(x, y, t)$ which is obtained by piecewise linear interpolation in time of the function $100 \exp{\Bigl( -0.8\bigl((x-l/2)^2 +(y-l/2)^2)\bigr)\Bigr)}\delta(t-0.2)$, where $\delta$ is the discrete-time unit impulse.

We choose a uniform grid in space. 
The gradients are discretized with second-order finite differences at each grid point $(i, j)$ as $\frac{\partial T}{\partial x}\big|_{i,j} \approx \frac{T_{i+1,j} - T_{i-1,j}}{2h}$, $\frac{\partial T}{\partial y}\big|_{i,j} \approx \frac{T_{i,j+1} - T_{i,j-1}}{2h}$ and the Laplacian is discretized as $(\frac{\partial^2 T}{\partial x^2} + \frac{\partial^2 T}{\partial y^2})\big|_{i,j} \approx \frac{T_{i+1, j} + T_{i-1, j} + T_{i, j+1} + T_{i, j-1} - 4T_{i, j}}{h^2}$. 
The spatially discretized form of equation \eqref{eq:conv_diff} is an LTI system, where $ T$ represents the state variable. The spatial discretization scheme dictates the sparsity pattern and the elements of the state matrix $\Tilde{A} \in \mathbb{R}^{400 \times 400}$. The other state-space matrices are $\Tilde{B} = \mathcal{I}_{400}$, $\Tilde{C} = \mathcal{I}_{400}$ and $\Tilde{D} = 0$. 

\begin{figure}
    \begin{minipage}[b]{0.48\textwidth}
        \centering
        \vspace{0pt}
        \includegraphics[width=0.9\textwidth]{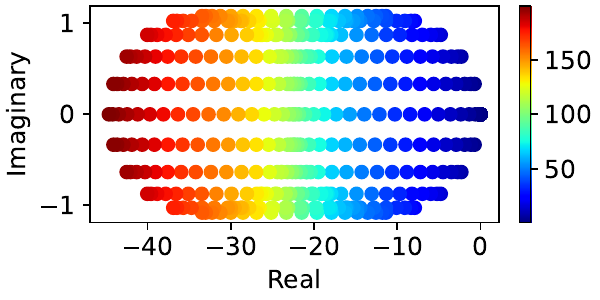}
        \captionof{figure}{Eigenvalues of the state matrix. The color bar shows eigenvalue clusters (Ex SM2.2).}
        \label{fig:conv_diff_evals}
    \end{minipage}\hfill
    \begin{minipage}[b]{0.48\textwidth}
        \centering
        \vspace{0pt}
        \includegraphics[width=0.6\textwidth]{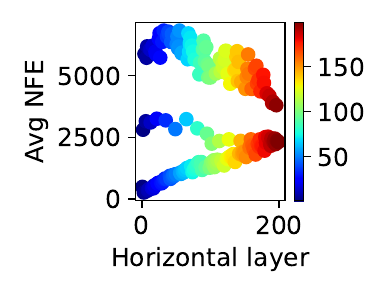}
        \captionof{figure}{Average NFE count of horizontal layers. The color bar shows horizontal layers (Ex SM2.2).}
        \label{fig:conv_diff_nfe}
    \end{minipage}
\vspace{-1.8em}
\end{figure}

\begin{figure}[htbp]
\begin{minipage}[t]{0.35\textwidth}
        \centering
        \includegraphics[width=\textwidth]{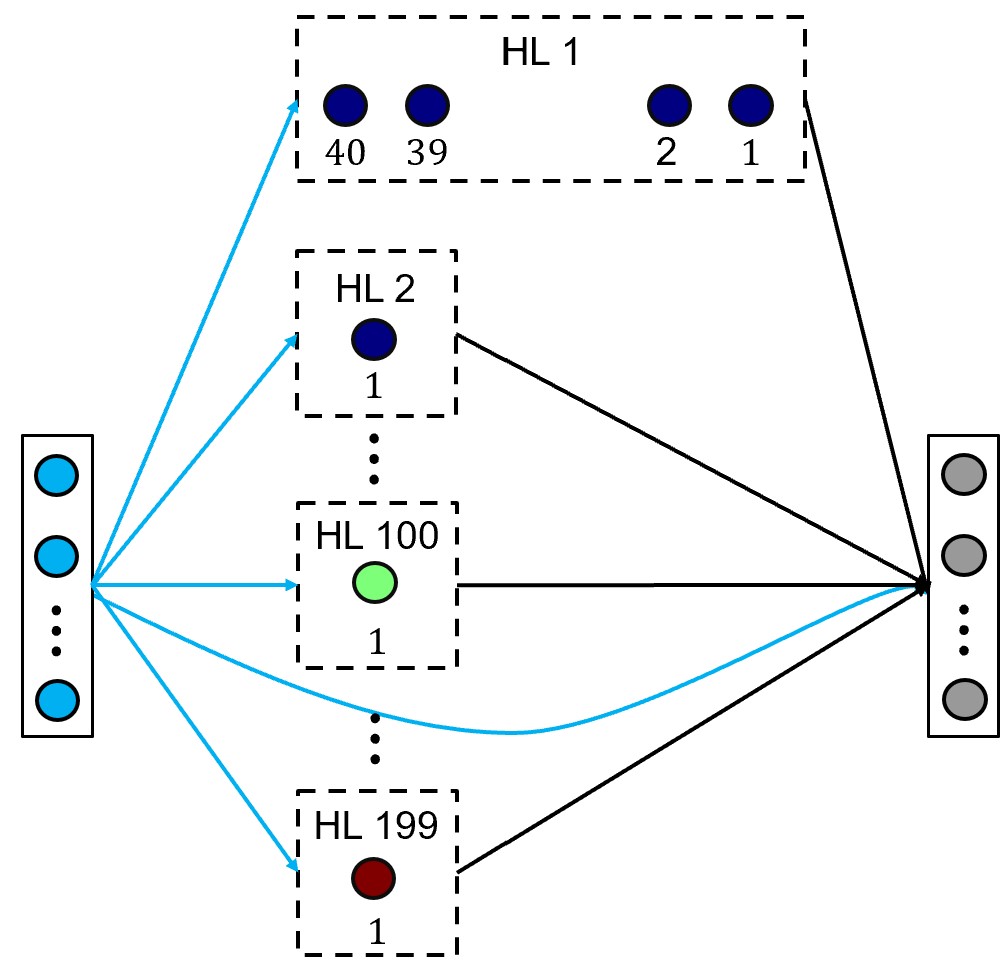}
        \captionof{figure}{DyNN architecture. Horizontal Layers (HLs) are indicated by the color bar in \cref{fig:conv_diff_evals}) (Ex SM2.2).}
        \label{fig:conv_diff_DyNN_architecture}
    \end{minipage}\hfill
    \begin{minipage}[t]{0.6\textwidth}
        \centering
        \includegraphics[width=0.85\textwidth]{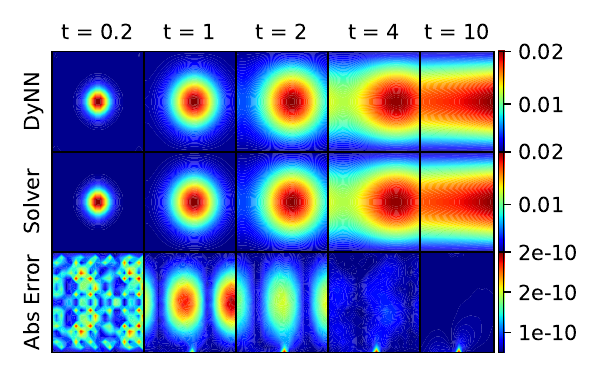}   
        \captionof{figure}{Convection diffusion equation. Top Panel: DyNN solution. Middle panel: numerical solution. Bottom panel: absolute error between the two solutions at five time instants (Ex SM2.2).}
        \label{fig:conv_diff_comparison}
    \end{minipage}
\vspace{-1.8em}
\end{figure}

We simulate the semi-discretized LTI system with a DyNN and compare the results with ones obtained from the classical numerical solver that simulates the LTI system using the Python routine \texttt{Scipy.signal.lsim}.
After preprocessing the LTI system with \cref{alg:preprocessing_lti}, the state matrix is block-diagonalized. 

\cref{fig:conv_diff_evals} shows the eigenvalue clustering. 
The state matrix, in this case, is not normal and hence not unitarily diagonalizable. 
The state matrix has 162 pairs of complex eigenvalues and 36 real eigenvalues with an algebraic multiplicity of 1 and has a repeated eigenvalue - $0$ with an algebraic multiplicity of 40. 
Thus, the repeated eigenvalues are clustered together, resulting in one horizontal layer with 40 neurons. \cref{fig:conv_diff_DyNN_architecture} shows the resulting architecture, where the first HHL has 40 neurons, and the rest have 1 neuron each. 
Although not apparent from the \cref{fig:conv_diff_DyNN_architecture}, the DyNN has 162 second-order neurons. 
The ones shown in the figure are all first-order neurons. 
The condition number of the transformation matrix, with 199 clusters, is $C_{tr} = 7.42$. 
If the number of clusters is increased by $1$, the repeated eigenvalues are forced to be in different clusters, which is unrealistic. Hence, the number of clusters should not be increased further. 

Due to the partial decoupling of state dynamics across the diagonal blocks, the NFE count of ODE solvers for neurons in any horizontal layer is independent of the NFE count of neurons in other horizontal layers.
\cref{fig:conv_diff_nfe} illustrates a high variation in the NFE count of neurons averaged over each horizontal layer. 
Finally, \cref{fig:conv_diff_comparison} demonstrates that the DyNN simulates the semi-discretized convection-diffusion system accurately up to machine precision compared to the numerical solver.

\begin{remark}
The right boundaries are not included as state variables in the finite difference discretization, as they are the same as the left boundary points. However, the top and bottom boundaries which are fixed, are included as the state variables, where the gradient with respect to time does not change. For convenience, the domain sizes are adapted to keep the discretization width $h_x = h_y = 0.5$.   
\end{remark}

\subsection{Another algorithm for the forward pass of a dynamic neural network}
\label{sm:timesteps_first}
\begin{figure}[htbp]
    \centering
    \includegraphics[width=0.6\textwidth]{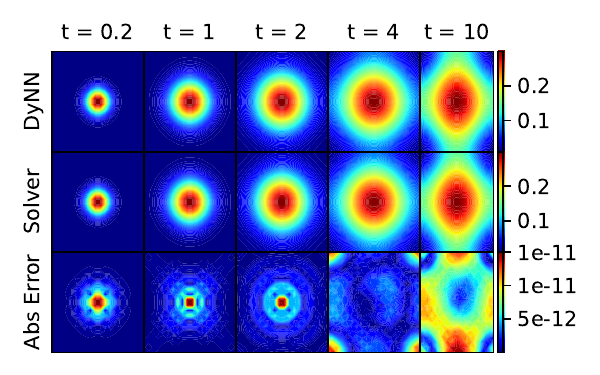}   
    \captionof{figure}{Example 3.1 Diffusion equation (see \cref{ss:diffusion} in the main text). Top Panel: DyNN solution with \cref{alg:dynamic neural network_timesteps}. Middle panel: numerical solution. Bottom panel: absolute error between the two solutions at five time instants.}
    \label{fig:diffusion_solution_timesteps}
\end{figure}

\begin{figure}[htbp]
    \centering
    \includegraphics[width=0.8\textwidth]{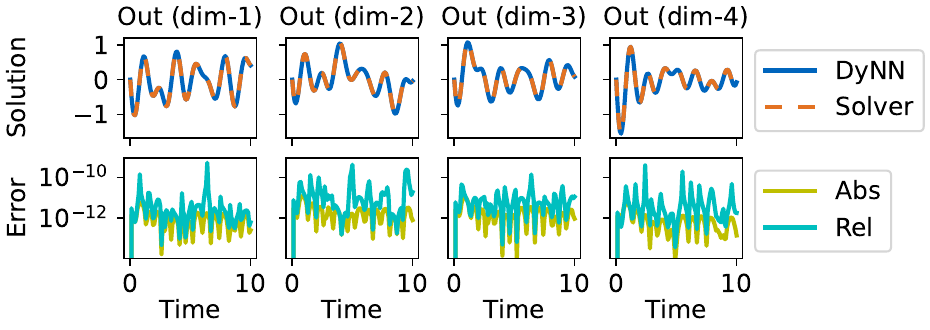}   
    \captionof{figure}{Example 3.2: The reason for horizontal layers (See \cref{ss:horizontal_layers} in the main text).
    Top panel: Outputs of DyNN with \cref{alg:dynamic neural network_timesteps} and numerical solver in all output dimensions. Bottom panel: relative and absolute errors between the DyNN output and numerical solution.}
    \label{fig:sol_sparsity_cond_timesteps}
\end{figure}

\begin{figure}[htbp]
    \centering
    \includegraphics[width=0.8\textwidth]{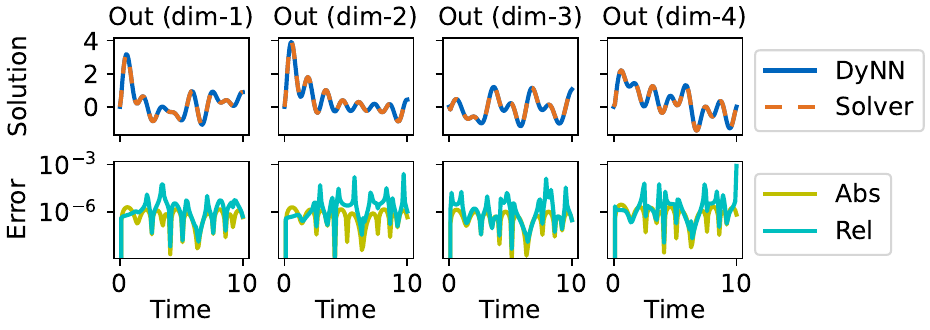}   
    \captionof{figure}{Example 3.3: State matrix with different kinds of clusters of eigenvalues (see \cref{ss:blobs} in the main text).  Top panel: Outputs of DyNN with \cref{alg:dynamic neural network_timesteps} and numerical solver in all output dimensions. Bottom panel: relative and absolute errors between the DyNN output and numerical solution.}
    \label{fig:hor_layers_timesteps}
\end{figure}

In this section, we present a slightly modified version of the \cref{alg:dynamic neural network} for performing a forward pass of a dynamic neural network. 
The key difference is that in \cref{alg:dynamic neural network}, the ODE corresponding to each neuron in the hidden layer is solved over the entire time domain, and a single interpolation function is used to compute states at intermediate points. 
In the \cref{alg:dynamic neural network_timesteps}, the ODE corresponding to each neuron in the hidden layer is solved over small intervals of times, which results in another outer loop over the time steps (See line 5 of \cref{alg:dynamic neural network_timesteps}). 
In this case, the number of ODE solves increases, but the time domain of each ODE is smaller.  
On each small time interval, we set the new initial conditions for the states as the final states of the previous time step.  
We set the parameter \texttt{dense\_output} to true as before. 
In \cref{alg:dynamic neural network_timesteps}, we fit one interpolation function per small time interval (instead of over the entire time domain). 
As a result, the interpolation error is reduced. 
We compute solutions of the same LTI systems considered in the numerical examples in the main text with identical problem setups but with \cref{alg:dynamic neural network_timesteps} instead.  
Figures \ref{fig:diffusion_solution_timesteps}, \ref{fig:sol_sparsity_cond_timesteps}, and \ref{fig:hor_layers_timesteps} show the respective solutions of the three LTI systems using constructed DyNNs along with the solutions with numerical solvers using \texttt{Scipy.signal.lsim} routine. 
We observe that performing a forward pass with \cref{alg:dynamic neural network_timesteps} results in errors that are lower in magnitude as compared to those obtained with \cref{alg:dynamic neural network} (See Figures \ref{fig:n_diffusion_solution}, \ref{fig:sol_sparsity_cond}, and \ref{fig:n_sol_error_blobs} for comparison). 
In particular, the errors are roughly 1 and 4 orders of magnitude lower for the first and second examples with \cref{alg:dynamic neural network_timesteps}. 

\renewcommand{\COMMENT}[2][.17\linewidth]{%
  \leavevmode\hfill\makebox[#1][l]{//~#2}}
    \begin{algorithm}[htbp]
    \caption{Forward pass of a dynamic neural network (ODEs solved over small time-intervals instead of over the entire time domain)}\label{alg:dynamic neural network_timesteps}
    \textbf{Input:} DyNN architecture and parameters - $\left(\mathcal{M},\mathcal{C},\mathcal{K},\mathcal{W},\Theta\right) \in \mathcal{P}_{dynn}^{hidden}, \left(\Phi,\Psi\right) \in \mathcal{P}_{dynn}^{output}$, inputs $u$ and $\dot{u}$ as function handles, time domain $\Omega = [t_0, t_f]$\\
    \textbf{Output:} Output of the dynamic neural network $\hat{y}$ as a function handle\\
    \textbf{Parameters:} \texttt{rtol, atol, method}
    \begin{algorithmic}[1]
        \For{$l \gets 1$ to $L$}
            \State \texttt{properties} $\gets$ \texttt{method, rtol, atol, dense\_output}
            \State \texttt{weights} $\gets \left( m^{(l)}_i,c^{(l)}_i,k^{(l)}_i, w_i^{(l)}, \phi_i^{(l)}\right)$
            \State $t \gets t_0$
            \While{$t \le t_f$}
                    \For{$i \gets n_l$ to $1$} 
                        \If{$t = t_0$}
                            \State Set initial conditions $[\hat{y}_i^{(l)}]_{old}$ to $0$. 
                        \EndIf
                        \State $s = [t,t+\Delta t]$
                        \State $[\hat{u}_i^{(l)}]_{s}$ $\gets$ $\begin{bmatrix} [u^T]_{s}&[\dot{u}^T]_{s}&[(\hat{y}_{i+1}^{(l)})^T]_s&\cdots&[(\hat{y}_{n_l}^{(l)})^T]_s
                        \end{bmatrix}$ 
                        \State $[\hat{y}_i^{(l)}]_{s}$ $\gets$ \texttt{solve\_ivp}($[\hat{y}_i^{(l)}]_{old}$, $[\hat{u}_i^{(l)}]_{s}$, s, \texttt{weights}, \texttt{properties})   
                        \State $[\hat{y}_i^{(l)}]_{old}$ $\gets$ $[\hat{y}_i^{(l)}]_{s}(t + \Delta t)$.
                    \EndFor
                    \State $t \gets t + \Delta t$
            \EndWhile
        \EndFor
    \State Compute DyNN output $\hat{y}\gets\biggl(\sum_{l=1}^{L} \sum_{i=1}^{n_l} \phi_i^{(l)}  \, \hat{y}_i^{(l)} \biggr) + \Psi \, u$ 
    \end{algorithmic}
\end{algorithm}

